\documentclass{article}

\usepackage[preprint]{neurips_2026}

\usepackage[utf8]{inputenc} 
\usepackage[T1]{fontenc}    
\usepackage{hyperref}       
\usepackage{url}            
\usepackage{booktabs}       
\usepackage{amsfonts}       
\usepackage{nicefrac}       
\usepackage{microtype}      
\usepackage{xcolor}         
\usepackage{graphicx}
\usepackage{subcaption}
\usepackage{amsmath}
\usepackage{booktabs}
\usepackage{multirow}
\usepackage{listings}

\definecolor{codegreen}{rgb}{0,0.6,0}
\definecolor{codegray}{rgb}{0.5,0.5,0.5}
\definecolor{codepurple}{rgb}{0.58,0,0.82}
\definecolor{backcolour}{rgb}{0.95,0.95,0.92}

\lstdefinestyle{mystyle}
{
    backgroundcolor=\color{backcolour},    
    commentstyle=\color{codegreen},
    keywordstyle=\color{magenta},
    numberstyle=\tiny\color{codegray},
    stringstyle=\color{codepurple},
    basicstyle=\ttfamily\footnotesize, 
    breakatwhitespace=false,          
    breaklines=true,                  
    captionpos=t,                     
    keepspaces=true,                  
    numbers=left,                     
    numbersep=5pt,                   
    showspaces=false,                 
    showstringspaces=false,
    showtabs=false,                   
    tabsize=4,
    upquote=true,
    literate={"}{{\fontencoding{T1}\selectfont\char34}}1
             {~}{{\char126}}1
}
\newif\ifcomments
\commentstrue
\ifcomments
    \providecommand{\ion}[1]{{\color{blue}{/* ion: #1 */}}}
\else
    \providecommand{\ion}[1]{}
\fi

\title{Uncovering Intra-expert Activation Sparsity for Efficient Mixture-of-Expert Model Execution}

\author{
  Jongseok Park\thanks{Work done during an internship at Advanced Micro Devices, Inc.}\\
  University of California, Berkeley\\
  \texttt{js\_park@berkeley.edu} \\
  \And
  Sunga Kim\\
  University of California, Berkeley\\
  \texttt{sunga.kim@berkeley.edu} \\
  \And
  Zhenyu Gu\\
  Advanced Micro Devices, Inc.\\
  \texttt{Zhenyu.Gu@amd.com} \\
  \And
  Ion Stoica\\
  University of California, Berkeley\\
  \texttt{istoica@berkeley.edu} \\
  \And
  Alvin Cheung\\
  University of California, Berkeley\\
  \texttt{akcheung@berkeley.edu} \\
}

\begin{document}

\maketitle

\begin{abstract}
Mixture of Experts (MoE) architecture has become the standard for state-of-the-art large language models, owing to its computational efficiency through sparse expert activation. However, sparsity through finer expert granularity is becoming increasingly difficult to achieve due to fundamental training challenges such as expert collapse and load imbalance. In this work, we explore and leverage \textbf{intra-expert activation sparsity} as a complementary and underexplored dimension of sparsity in MoE models. Surprisingly, substantial intra-expert sparsity is \textit{readily available in existing pre-trained MoE models}, without any modification to the activation function or model parameters, providing up to \textbf{90\% sparsity} within each expert without significant accuracy loss. We explore intra-expert activation sparsity across eight off-the-shelf MoE models ranging from 1B to 400B parameters, and extend the MoE execution pipeline of vLLM to leverage intra-expert activation sparsity by skipping the computations of inactive neurons, on top of its existing optimizations, achieving up to 2.5$\times$ speedup in MoE layer execution and 1.2$\times$ end-to-end speedup compared to the original dense vLLM baseline.

\end{abstract}

\section{Introduction}
\label{sec:intro}

Mixture of Experts (MoE) architecture~\cite{shazeer2017outrageouslymoe, fedus2022switch} has become the de facto standard for state-of-the-art Large Language Models (LLMs)~\cite{qwen3.5, openai2025gptoss120bgptoss20bmodel, llama4_2026}, owing to its superior computational efficiency over traditional dense LLM architectures. 
MoE models use a sparse feed-forward network (FFN) composed of multiple independent sub-networks called \textit{experts}. 
Each expert typically specializes in a distinct domain of knowledge~\cite{zhang2022moefication}, and during inference, the model selectively activates only those experts most relevant to the current input token. 
Unlike dense models, which activate all parameters, this sparse activation allows MoE models to reduce computation while maintaining high model capacity. As a result, MoE models have evolved to increase the number of experts and maximize sparsity by reducing the percentage of active experts.

Training these sparse MoE models is notoriously difficult. Without careful regularization, MoE models are prone to expert collapse~\cite{fedus2022switch, zoph2022stmoe}, where the model converges to using a few general experts while leaving the majority of experts untrained, or load imbalance~\cite{fedus2022switch, zhou2022expertchoice_loadimbal}, where hotspots in expert activation degrade both model quality and hardware utilization. Additionally, MoE models suffer from representation collapse where experts converge to similar specializations~\cite{chi2022representationcollapse} and reduced per-expert gradient updates~\cite{panda2025densebackprop_moe}. These challenges only compound as the number of experts grows, posing fundamental difficulties for further increasing \textit{inter-expert sparsity}.

Instead of pushing inter-expert sparsity further, we explore sparsity within the activations of each expert, or \textbf{intra-expert activation sparsity} as a complementary source of sparsity in MoE models at inference time. 
Activation sparsity has been widely studied in model architectures with ReLU activations that explicitly zero negative activations~\cite{glorot2011deep}, but recent studies reveal that activation sparsity exists in newer LLM architectures as well~\cite{szatkowski2025universalpropertiesactivationsparsity, cheng2025mixtureofneuronexp, lee2024cats, liu2024teal}, despite their use of leaky activation functions such as SwiGLU~\cite{shazeer2020swiglu}. Activation outputs of LLM models are known to follow a Sparsing Law~\cite{luo2024sparsinglaw}, where the fraction of important neurons decreases as the model parameter size grows. For example, activation outputs of models such as Llama-2~\cite{touvron2023llama2} that use SwiGLU follow a long-tailed distribution with longer tails in the larger variants of their model family~\cite{szatkowski2025universalpropertiesactivationsparsity}. The same phenomenon has also been validated on MoE models~\cite{cheng2025mixtureofneuronexp} such as Qwen3~\cite{yang2025qwen3technicalreport} and Mixtral~\cite{jiang2024mixtral}, confirming that activation sparsity is not an optimization axis restricted only to ReLU-based LLMs. 

\begin{figure}[t]
    \centering
    \hspace*{-18pt}
    \includegraphics[width=0.88\linewidth]{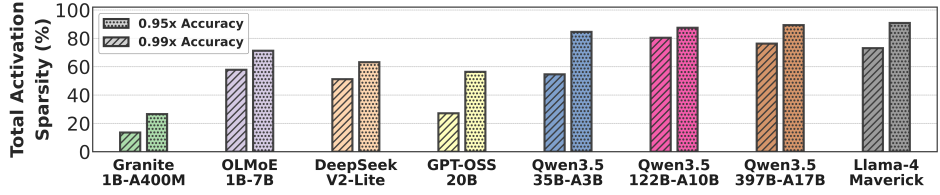}
    \caption{Maximum total intra-expert activation sparsity that retains 95\% and 99\% of the average benchmark score of the unmodified execution. Benchmark details are available in~\ref{sec:intra-expert-meth}.}
    \label{fig:sparsity_accuracy}
    \vspace{-12pt}
\end{figure}


However, we find \textbf{two major obstacles} that inhibit the widespread use of intra-expert activation sparsity for efficient MoE inference. First, many existing works target older dense architectures or require training to achieve sparsity. This greatly limits the scope and practicality of the technique, calling into question whether the approach is applicable to more widely used \textit{pretrained MoE models}.
Secondly, there is no \textit{end-to-end integration} of intra-expert activation sparsity in highly optimized execution systems. Production systems such as vLLM~\cite{kwon2023efficientmemorymanagementlarge} include sophisticated MoE kernels and techniques such as expert parallelism~\cite{deepseekv2_2024, lepikhin2021gshard} to maximize performance. However, existing works only implement a naive research-level baseline, leaving a critical gap in evaluating the compatibility and benefit of intra-expert activation sparsity in high-performance systems.

To this end, we first conduct a thorough study of intra-expert activation sparsity on off-the-shelf MoE models ranging from 1B to 400B parameters and report a surprising finding: substantial intra-expert activation sparsity is \textit{readily available} in many existing MoE models, without requiring any modification to the model. By ranking neurons by their activation scores and deactivating the lowest-scoring ones, we measure up to \textbf{90\% sparsity} within active experts while retaining 95\% of the original model accuracy, as shown in Figure~\ref{fig:sparsity_accuracy}. We also find that neurons from shared and routed experts have markedly different impacts on accuracy, to the point when all shared expert neurons are active, as few as 2\% of the active routed expert parameters suffice to maintain model capacity for large models such as Qwen3.5-122B-A10B.

Based on this study, we extend the fused MoE execution pipeline of the popular production-grade inference system vLLM to leverage intra-expert activation sparsity by skipping the up and down projection computations for inactive neurons. Our implementation builds on top of the expert-batched MoE kernel design of vLLM, retaining \textbf{full support for existing optimizations} such as HIP Graph~\cite{amd2024hipgraphs} and CUDA Graph~\cite{nvidia2019cudagraphs}, kernel fusion~\cite{amd2024miopen_fusion}, and expert parallelism. We also share the insights we gained during our implementation to clearly illustrate the benefits and limitations of intra-expert sparsity in an end-to-end production setting.

In summary, we make the following contributions. (1) We explore intra-expert activation sparsity across eight off-the-shelf MoE models, with thorough analysis on their activation sparsity characteristics. (2) We redesign the MoE execution pipeline of vLLM to efficiently exploit both inter- and intra-expert sparsity on GPUs. (3) We integrate our sparse execution pipeline into the full vLLM engine and demonstrate performance gains, achieving up to 2.5$\times$ speedup in MoE layer execution and 1.2$\times$ speedup in end-to-end inference. Together, these contributions establish intra-expert activation sparsity as a practical optimization technique for MoE model execution and lay the groundwork for future efficient LLM research.

\section{Background and Related Work}
\label{sec:background}

\noindent{\textbf{Feed-Forward Network: }}
The feed-forward network (FFN) of the Transformer architecture~\cite{vaswani2017attention} applies a non-linear activation function in a higher-dimensional latent space to encode and process factual and linguistic knowledge~\cite{geva2021transformer}, and often dominates the parameter count of LLMs. Each dimension of the intermediate latent space of an FFN corresponds to an individual neuron whose contribution to the output is weighted by its activation value. Early work has shown that FFN neurons specialize in distinct aspects of the input, and a large fraction of neurons return near-zero activation values, effectively becoming inactive for that input~\cite{zhang2022moefication, mirzadeh2023relustrikesback}. 

\noindent{\textbf{Mixture of Experts: }}
Mixture of Experts (MoE) FFN architecture leverages the specialization of neurons by replacing the single FFN with multiple independent expert FFNs, where each expert learns to specialize on a distinct subset of inputs~\cite{shazeer2017outrageouslymoe, fedus2022switch}. During inference, a router network determines which experts should be activated for a given input token, and the vast majority of FFN neurons belonging to unrelated experts are efficiently bypassed, providing coarse-grained \textit{inter-expert sparsity}. However, training such MoE models is known to be extremely difficult~\cite{fedus2022switch, zoph2022stmoe, chi2022representationcollapse}, as the model must simultaneously specialize each expert toward distinct domains while ensuring equal importance of the domains for balanced expert utilization. Since expert specialization must also emerge \textit{implicitly} during training rather than being explicitly prescribed, MoE training is highly unstable without careful regularization, making models prone to expert collapse~\cite{zoph2022stmoe} or load imbalance~\cite{zhou2022expertchoice_loadimbal}.

\noindent{\textbf{Activation Sparsity: }}
Sparsity in neural networks can also be achieved through activation sparsity. Unlike inter-expert sparsity, which is built into the MoE model architecture, activation sparsity exploits the tendency of \textit{activation functions} to suppress a large fraction of neuron outputs for any given input. For example, the hard zero threshold of ReLU activation causes a significant fraction of neurons to be completely inactive~\cite{glorot2011deep}, allowing computations to be skipped entirely. While activation sparsity was pioneered in the context of convolutional neural networks (CNNs) that used ReLU extensively~\cite{krizhevsky2012alexnet, simonyan2014vgg}, works such as DejaVu~\cite{liu2023dejavu} demonstrated the potential of activation sparsity for efficient LLM inference using ReLU-activated LLMs, such as OPT~\cite{zhang2022opt}, and have shown over 90\% sparsity~\cite{mirzadeh2023relustrikesback} in its FFN layers.

Recent studies confirm the existence of activation sparsity in models with leaky GeLU~\cite{hendrycks2016gelu} or SiLU~\cite{elfwing2017silu} based Gated Linear Unit (GLU) activations such as GEGLU or SwiGLU~\cite{shazeer2020swiglu}, despite their lack of exact zeros~\cite{szatkowski2025universalpropertiesactivationsparsity, cheng2025mixtureofneuronexp, lee2024cats, liu2024teal}. Sparsing Law~\cite{luo2024sparsinglaw} establishes that the fraction of activated neurons decreases as the model parameter size increases, regardless of the activation used. This observation is empirically shown to be true in LLMs using leaky activations~\cite{szatkowski2025universalpropertiesactivationsparsity}, such as Gemma3~\cite{gemmateam2025gemma3} and Qwen 2.5~\cite{yang2024qwen25}, and also MoE models such as Qwen3-30B-A3B~\cite{yang2025qwen3technicalreport} and Mixtral-8x7B~\cite{jiang2024mixtral}. These works collectively establish that while leaky activations lack exact zeros, many FFN neurons remain unimportant and can be leveraged for activation sparsity through appropriate thresholding. 

\section{Intra-expert Activation Sparsity}
\label{sec:intra-expert}
\subsection{Methodology}
\label{sec:intra-expert-meth}

\begin{table*}[t]
  \centering
  \resizebox{\textwidth}{!}{
  \begin{tabular}{lccccc}
    \multicolumn{6}{l}{"Experts" and "Active Experts" refer to the number of experts and active experts \textit{excluding} shared experts.} \\
    \toprule
    Model Name & Params (Active / Total) & Experts & Active Experts & FFN Dim. & Shared Experts \\
    \midrule
    Granite-1B-A400M~\cite{granite_team_2024} & 400M / 1B & 32 & 8 & 512 & False \\
    OLMoE-1B-7B~\cite{muennighoff2024olmoe} & 1B / 7B & 64 & 8 & 1024 & False \\
    DeepSeek-V2-Lite~\cite{deepseekv2_2024} & 2.4B / 16B & 64 & 6 & 1408 & True \\
    GPT-OSS-20B~\cite{openai2025gptoss120bgptoss20bmodel} & 3B / 20B & 32 & 4 & 2880 & False \\
    Qwen3.5-35B-A3B~\cite{qwen3.5} & 3B / 35B & 256 & 8 & 512 & True \\
    Qwen3.5-122B-A10B~\cite{qwen3.5} & 10B / 122B & 256 & 8 & 1024 & True \\
    Qwen3.5-397B-A17B~\cite{qwen3.5} & 17B / 397B & 512 & 10 & 1024 & True \\
    Llama-4-Maverick~\cite{llama4_2026} & 17B / 400B & 128 & 1 & 8192 & True \\
    \bottomrule
    
  \end{tabular}
  }
  \vspace{-4pt}
  \caption{Architectural details of the tested MoE models}
  \vspace{-18pt}
  \label{tab:model_architectures_expanded}
\end{table*}

We hypothesize that when a token is routed to an expert, it only needs to activate a small subset of that expert's neurons, not the expert as a whole. We refer to this as \textbf{intra-expert activation sparsity}.
To validate this claim, we take pre-trained MoE LLMs and measure their benchmark scores at different levels of intra-expert activation sparsity, by sorting each expert's post-activation outputs and zeroing out the lowest-scoring neurons. We do not modify any other part of the model architecture or update the model parameters in any way. 

We test eight MoE models as shown in Table~\ref{tab:model_architectures_expanded} against five benchmarks, namely ARC-Challenge~\cite{clark2018arc}, ARC-Easy~\cite{clark2018arc}, HellaSwag~\cite{zellers2019hellaswag}, Winogrande~\cite{sakaguchi2019winogrande}, and TruthfulQA-mc2~\cite{lin2021truthfulqa}, to cover a wide variety of models and benchmarks. We use lm-eval-harness~\cite{eval-harness} with vLLM~\cite{kwon2023efficientmemorymanagementlarge} backend, and evaluate both with and without applying intra-expert sparsity to the shared experts for models with shared experts. We define the maximum sparsity at which the model retains 95\% of its baseline average benchmark score as the representative \textit{sparsity cutoff} value of the model. License details of the assets are available in Appendix~\ref{apn:licenses}.

\subsection{Observation and Analysis}
\label{sec:intra-expert-analysis}

\begin{figure*}[t]
  \centering
  \captionsetup[subfigure]{skip=4pt} 
  
  \includegraphics[width=0.995\textwidth]{figures/per_model_task_legend_row.png}
  
  \vspace{1pt}
  
  \begin{subfigure}[b]{0.245\textwidth}
    \includegraphics[width=\textwidth]{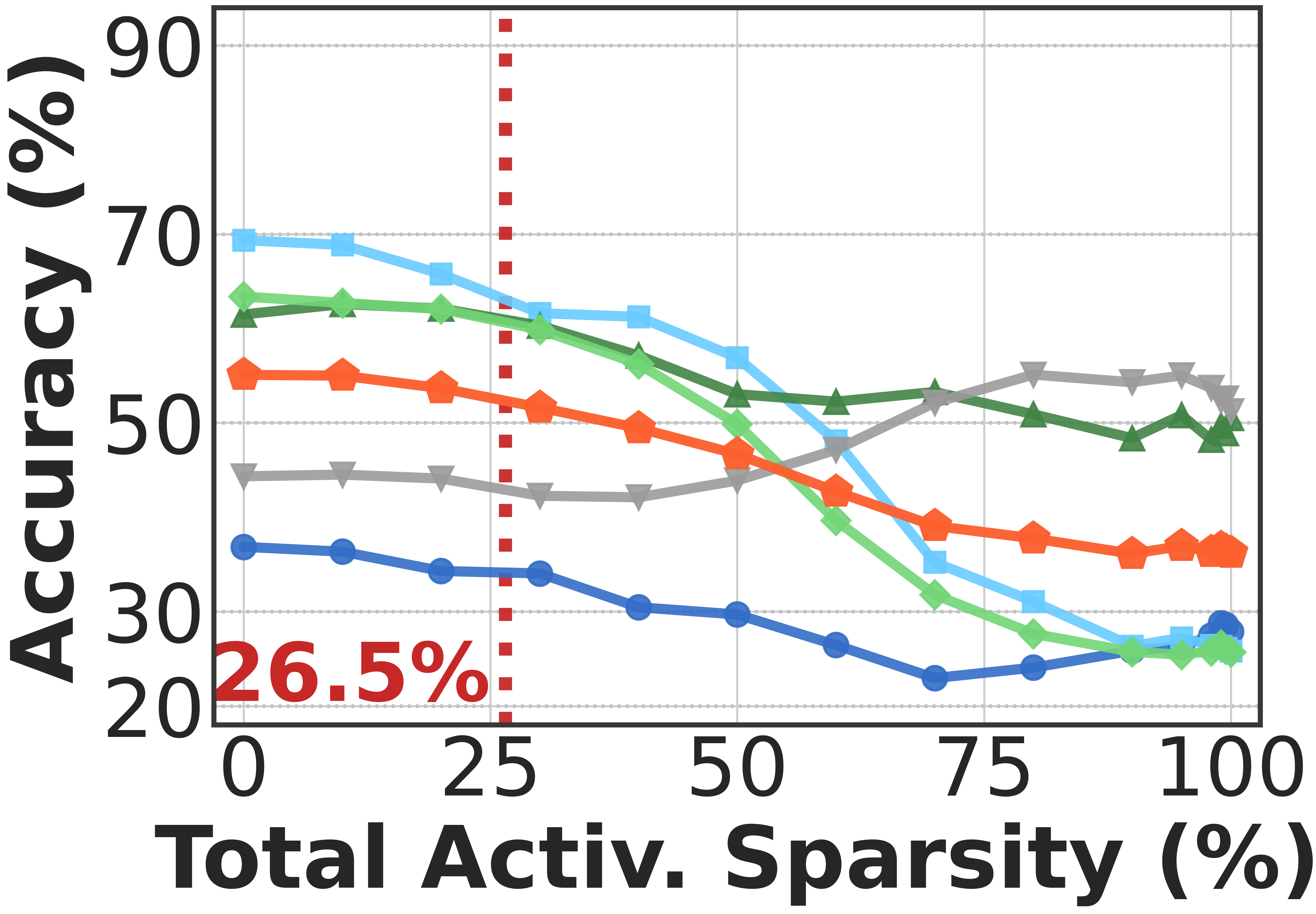}
    \caption{Granite-1B-A400M}
  \end{subfigure}\hfill
    \begin{subfigure}[b]{0.245\textwidth}
    \includegraphics[width=\textwidth]{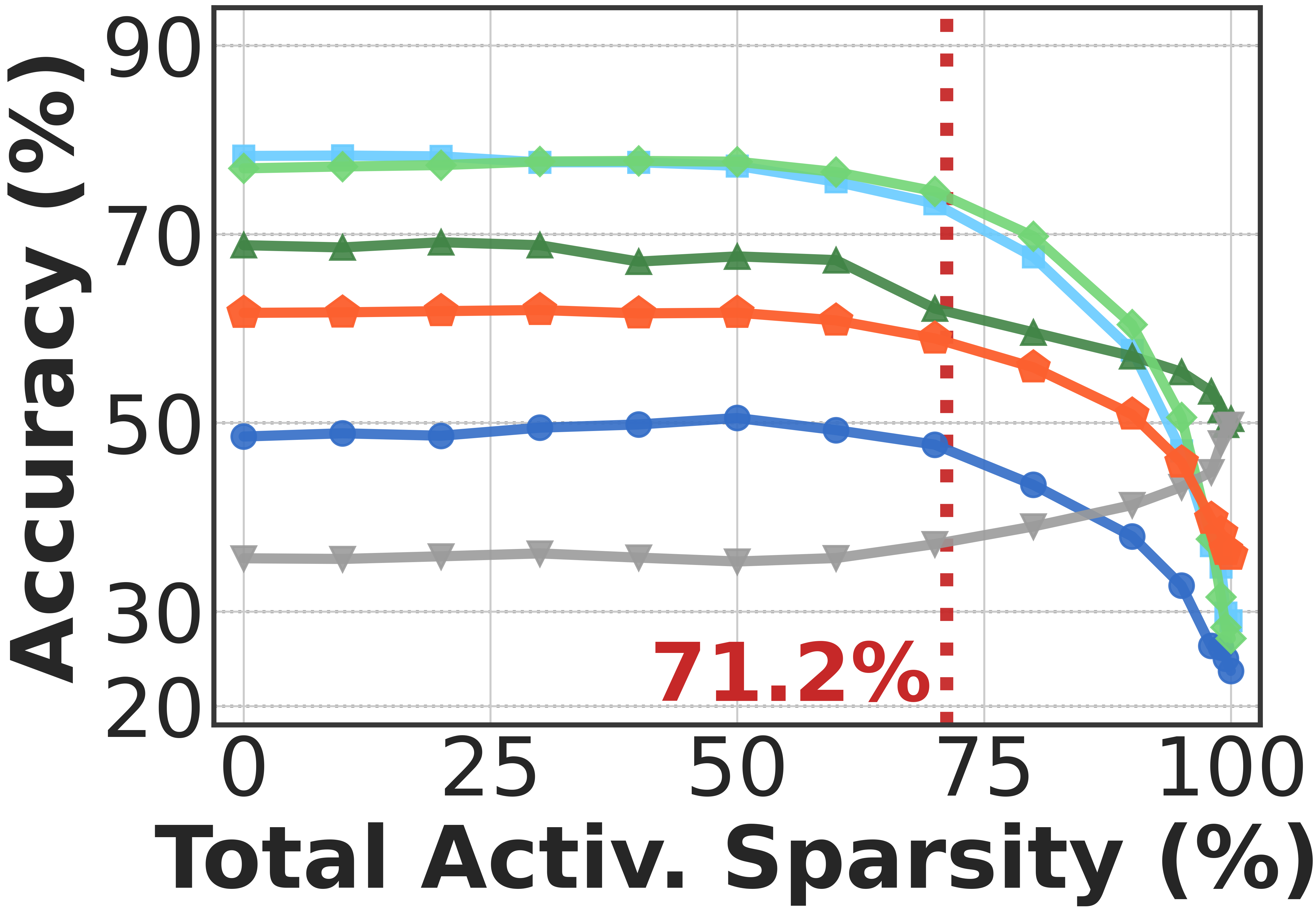}
    \caption{OLMoE-1B-7B}
  \end{subfigure}\hfill
  \begin{subfigure}[b]{0.245\textwidth}
    \includegraphics[width=\textwidth]{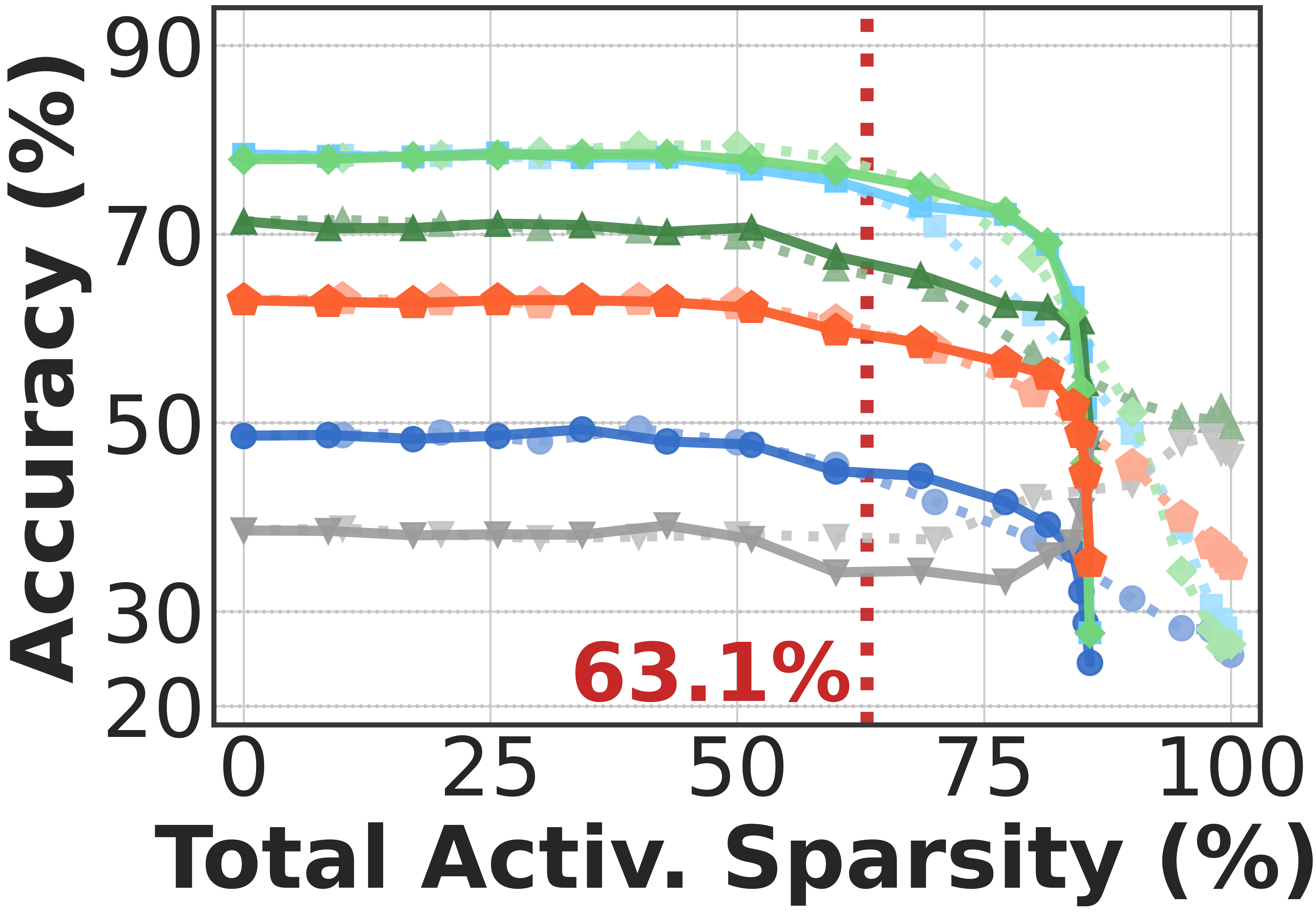}
    \caption{DeepSeek-V2-Lite}
    \label{fig:per_model_sparsity_accuracy_deepseekv2lite}
  \end{subfigure}\hfill
  \begin{subfigure}[b]{0.245\textwidth}
    \includegraphics[width=\textwidth]{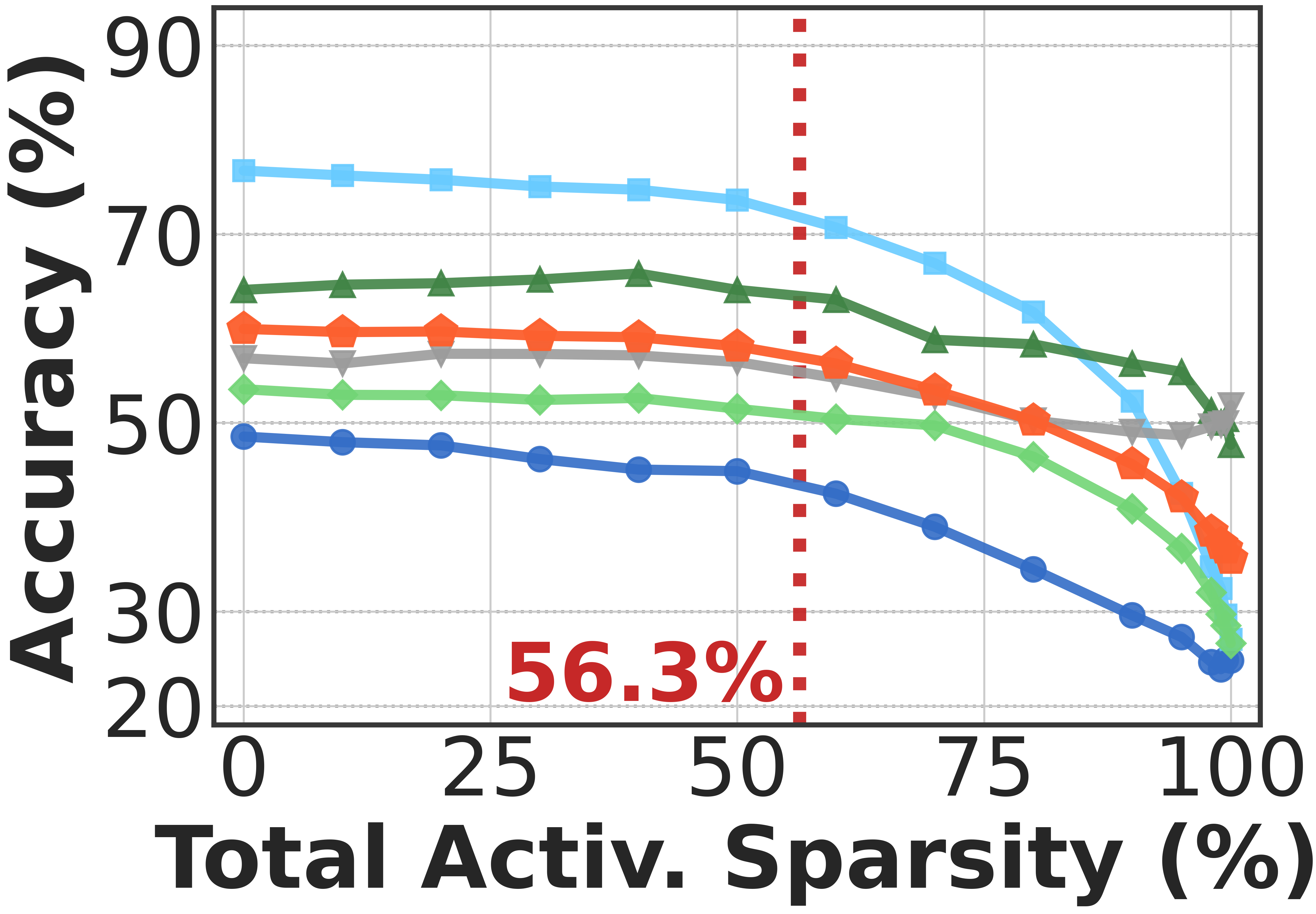}
    \caption{GPT-OSS-20B}
    \label{fig:per_model_sparsity_accuracy_gptoss_20b}
  \end{subfigure}

  \vspace{4pt}
  
  
  \begin{subfigure}[b]{0.245\textwidth}
    \includegraphics[width=\textwidth]{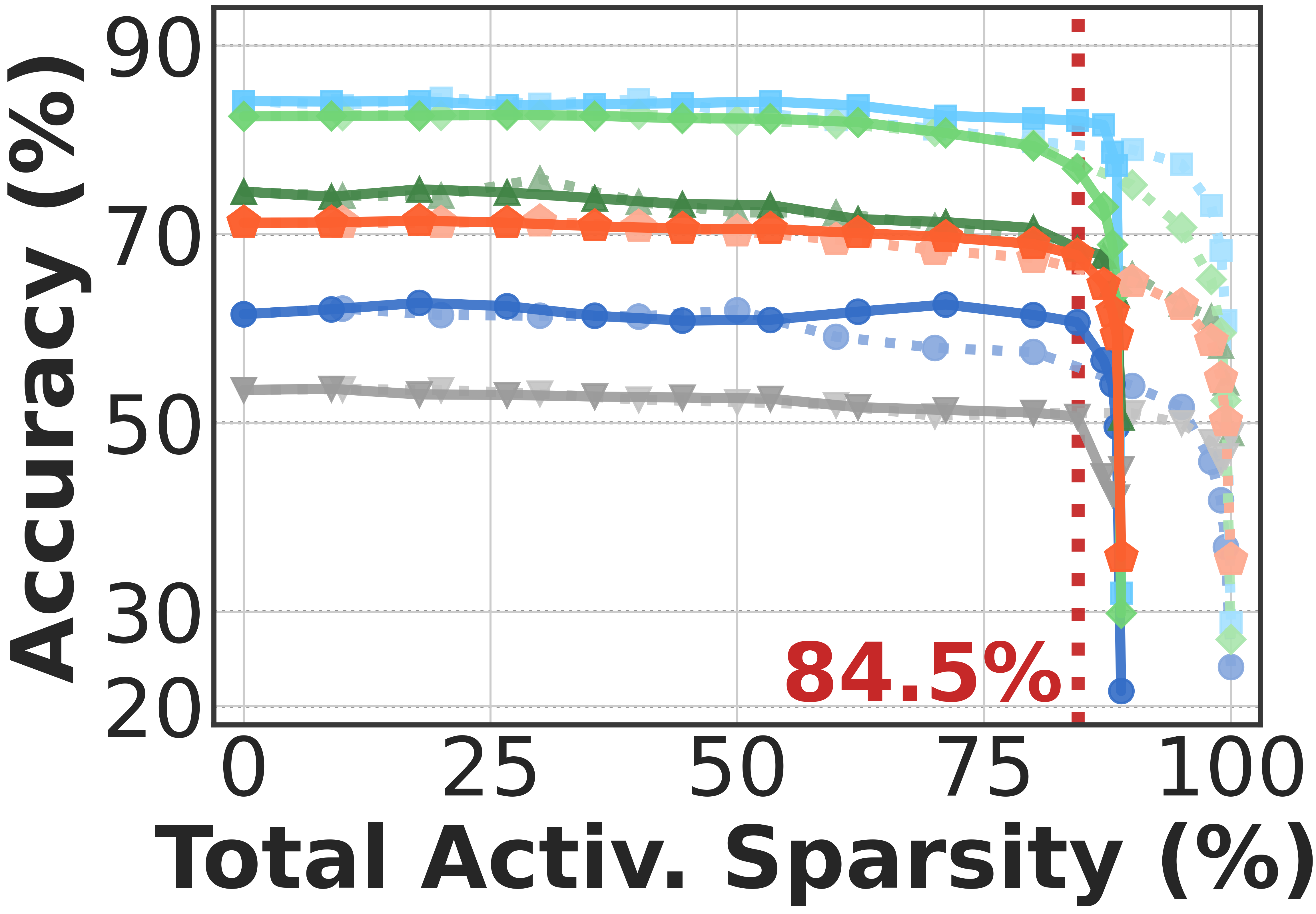}
    \caption{Qwen3.5-35B-A3B}
  \end{subfigure}\hfill
  \begin{subfigure}[b]{0.245\textwidth}
    \includegraphics[width=\textwidth]{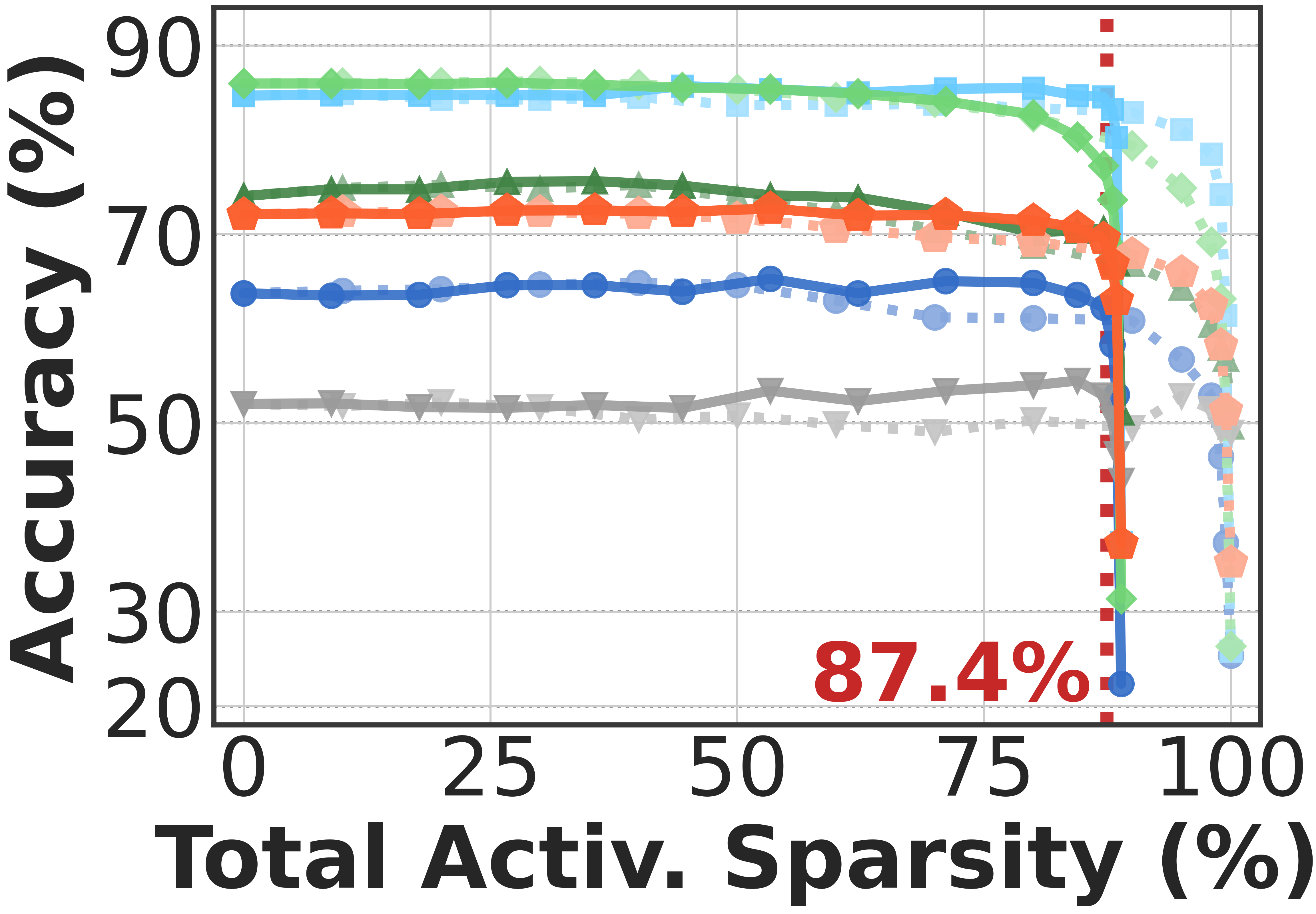}
    \caption{Qwen3.5-122B-A10B}
  \end{subfigure}
  \begin{subfigure}[b]{0.245\textwidth}
    \includegraphics[width=\textwidth]{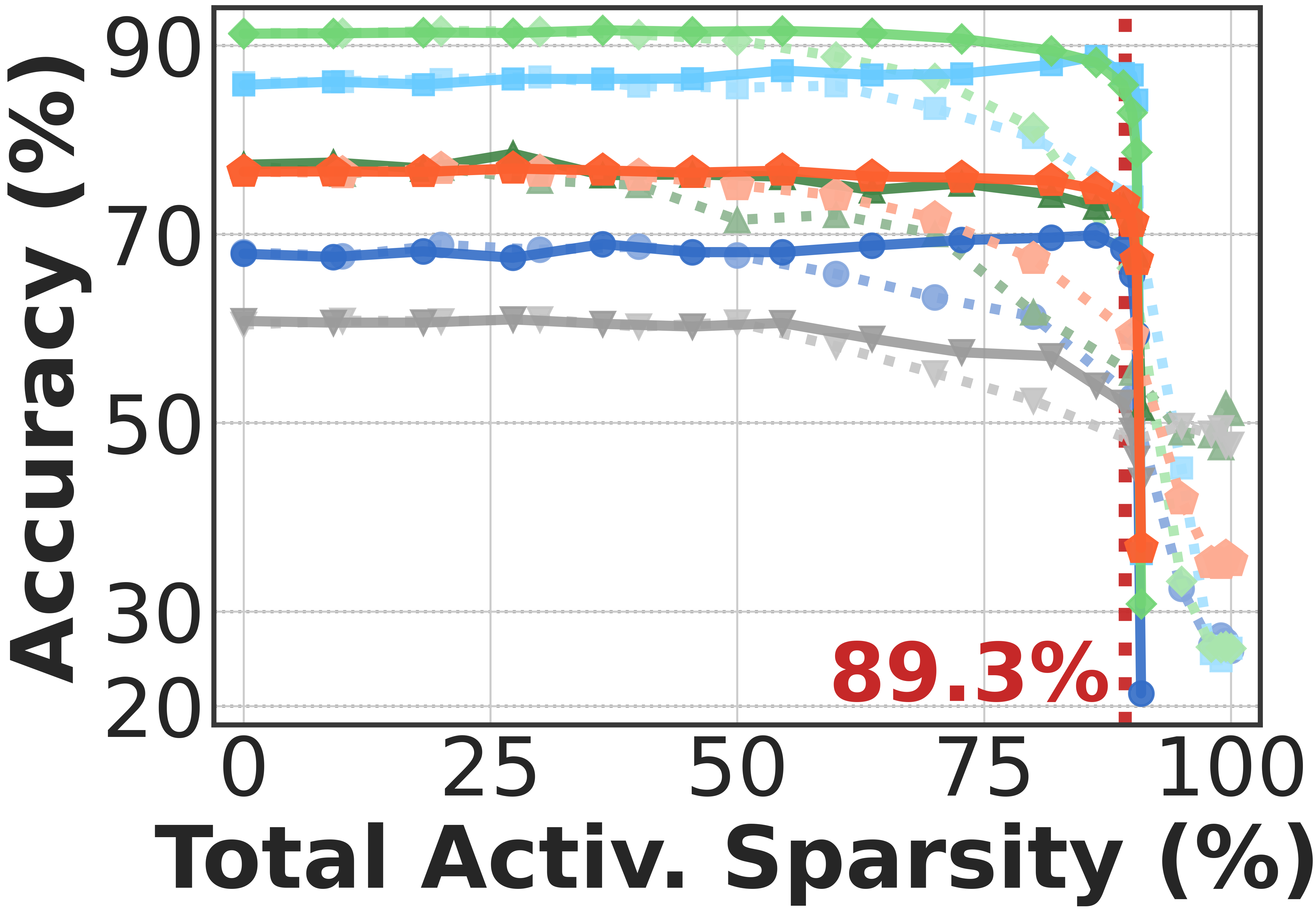}
    \caption{Qwen3.5-397B-A17B}
  \end{subfigure}\hfill
  \begin{subfigure}[b]{0.245\textwidth}
    \includegraphics[width=\textwidth]{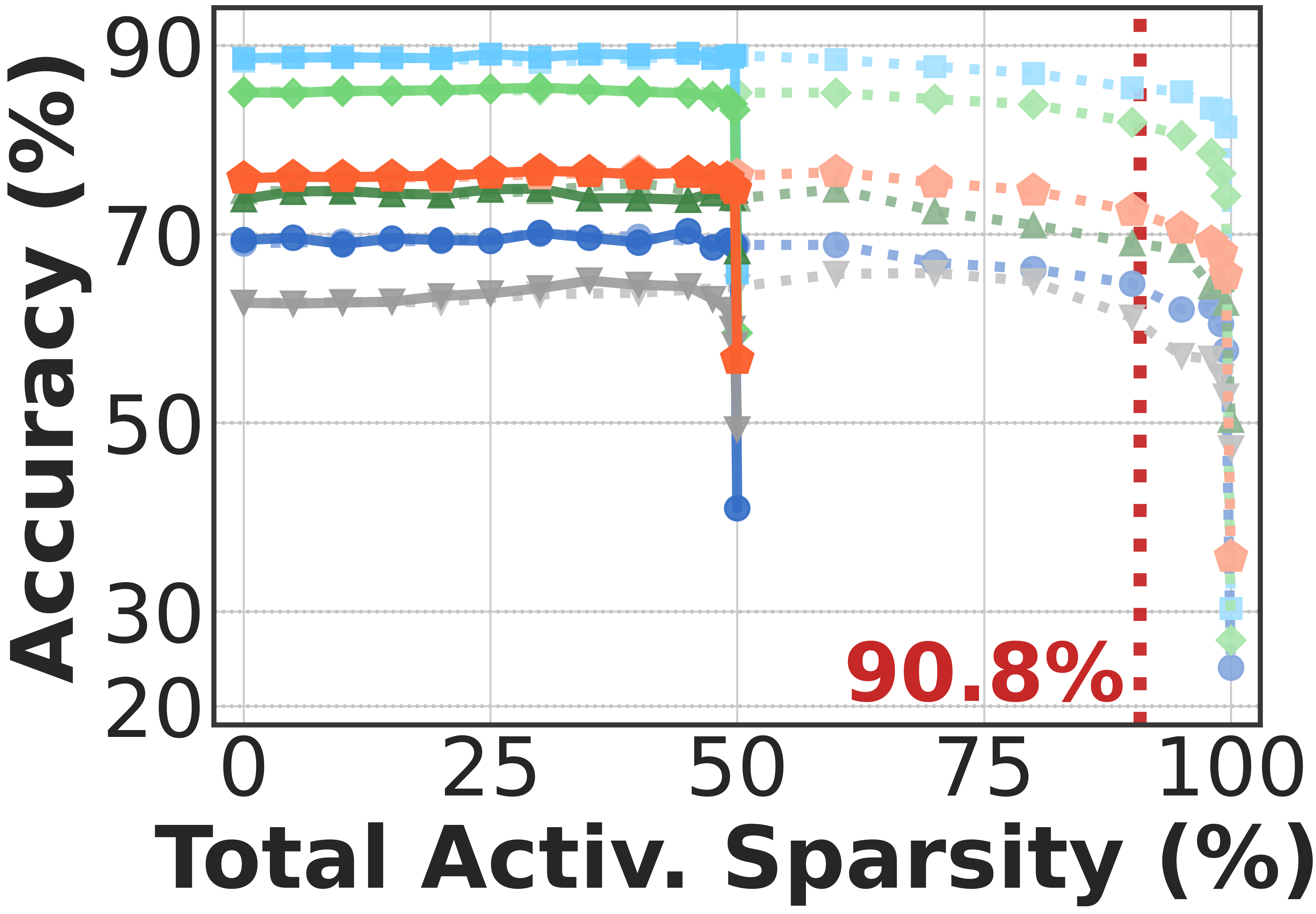}
    \caption{Llama-4-Maverick}
  \end{subfigure}\hfill
  \caption{Accuracy benchmark of MoE models with varying levels of total activation sparsity. Sparsity is applied only to routed experts in R results, and sparsity is applied to both routed and shared experts in R+S results. Sparsity cutoff of the model is annotated with a vertical red dotted line.}
  \label{fig:per_model_sparsity_accuracy}
  \vspace{-15pt}
\end{figure*}

We first present an overview and take a deeper dive comparing the importance of (1) neurons among the shared experts, (2) neurons across different routed experts, and (3) neurons within a single expert, to find which neurons are most influential for accuracy and guide the application of activation sparsity.

Figure~\ref{fig:per_model_sparsity_accuracy} presents the accuracy benchmark of the tested models. As noted in previous studies, we observe the trend that larger models generally have increased robustness to activation sparsity, with the smallest Granite-1B-A400M having the lowest sparsity cutoff at 26.5\% and the largest Llama-4-Maverick having the highest cutoff at 90.8\%. However, unlike previous studies that only covered models up to 30B parameters~\cite{szatkowski2025universalpropertiesactivationsparsity, cheng2025mixtureofneuronexp}, our experiment extends the model size up to 400B and confirms that intra-expert activation sparsity scales well beyond models with 100B parameters with even higher sparsity levels. From this observation, we conclude that intra-expert activation sparsity is not only readily available in existing models, but also \textbf{increasingly useful} in larger frontier models.

\noindent{\textbf{Shared Expert Neurons:}}
We find that shared experts have higher per-neuron impact on accuracy compared to neurons of routed experts. For example, in Figure~\ref{fig:per_model_sparsity_accuracy_deepseekv2lite}, DeepSeek-V2-Lite is shown to retain its accuracy over a wider range of sparsity in its routed expert-only (R) results compared to GPT-OSS-20B in Figure~\ref{fig:per_model_sparsity_accuracy_gptoss_20b}, despite the smaller parameter count of 16 billion. However, in the dotted (R+S) results where the sparsity is also applied to the shared experts, its accuracy curve becomes similar to GPT-OSS-20B with a wider arc and reduced accuracy retention. Similar trends are also observed in other models with shared experts, where the (R) results show a much sharper knee and constantly higher accuracy against the (R+S) results across all sparsity.

This phenomenon is more pronounced in larger parameter models, to the point where \textbf{less than 2\% of the active routed expert neurons is enough to retain model capacity} for each token when combined with the full shared expert, as shown by the sparsity cutoff of the Qwen3.5-122B-A10B and Qwen3.5-397B-A17B models. 
While prior work~\cite{qwen3.5, deepseekv2_2024} has pursued higher sparsity through inter-expert sparsity using finer expert granularity, achieving an equivalent 2\% activation level via inter-expert sparsity alone would require scaling the total number of experts by approximately \textit{50$\times$}, which would significantly worsen the training challenges such as expert collapse and load imbalance discussed in Section~\ref{sec:background}.

\begin{figure}[t]
  \centering
  \captionsetup[subfigure]{skip=4pt} 
  
  \begin{minipage}[t]{0.49\textwidth}
    \begin{subfigure}[t]{0.48\linewidth}
      \vspace{0pt}
      \includegraphics[width=\textwidth]{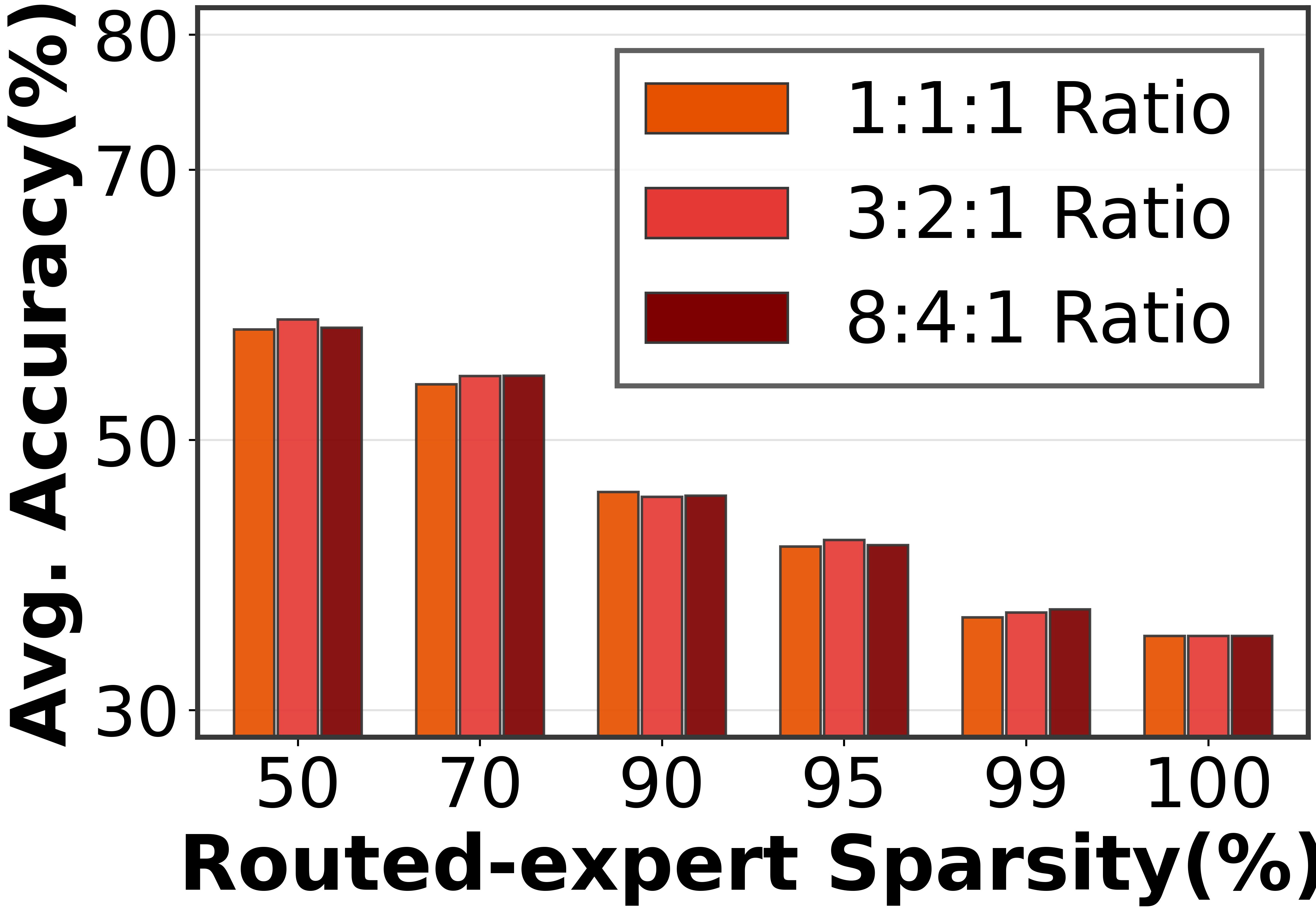}
      \caption{GPT-OSS-20B}
    \end{subfigure}\hfill
    \begin{subfigure}[t]{0.48\linewidth}
      \vspace{0pt}
      \includegraphics[width=\textwidth]{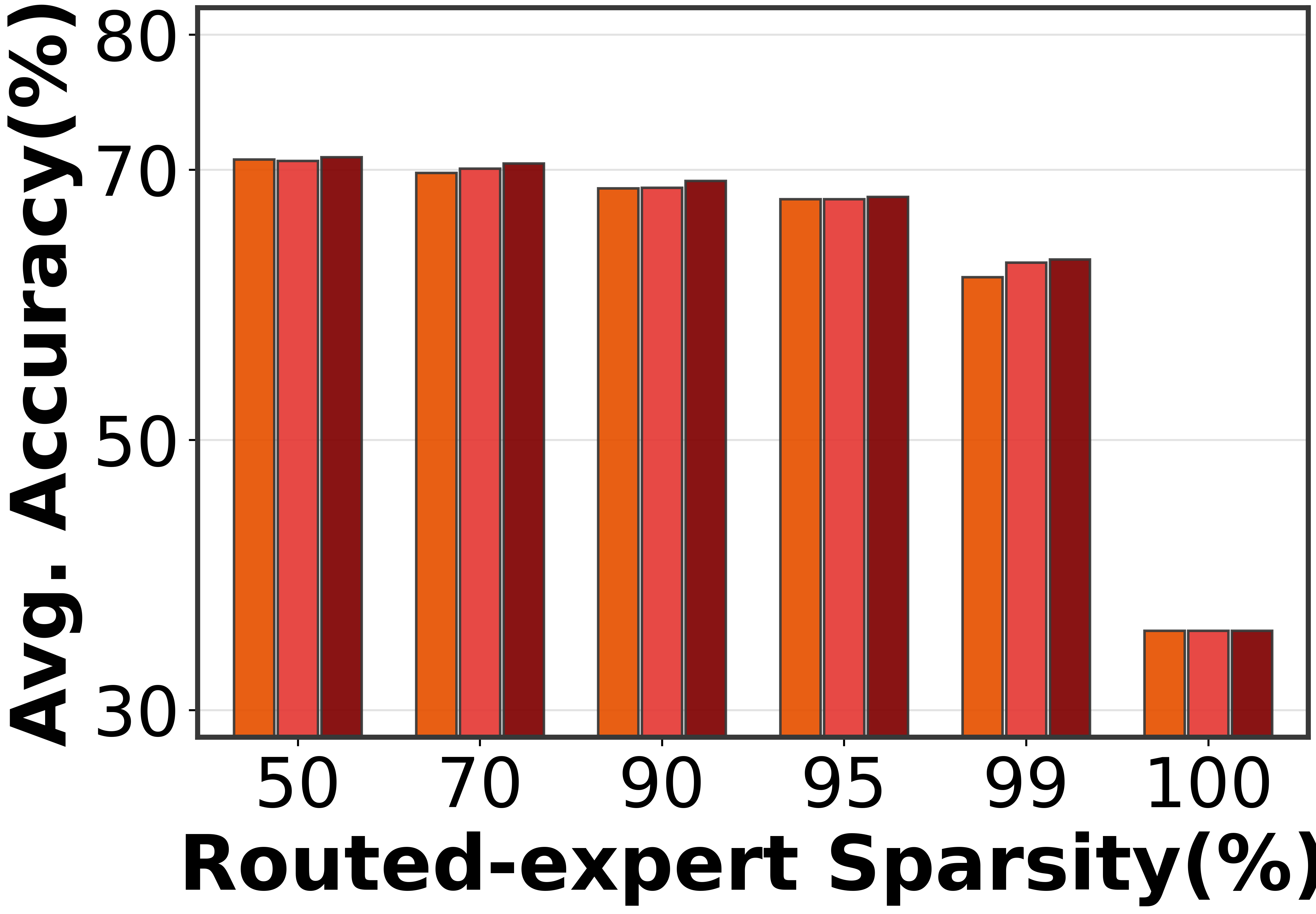}
      \caption{Qwen3.5-35B-A3B}
    \end{subfigure}
    \caption{Average accuracy when different ratio of neurons are allocated per expert based on router weight. Legend is shared between subfigures.}
    \label{fig:weighted_model_sparsity_accuracy}
  \end{minipage}\hfill
  \begin{minipage}[t]{0.49\textwidth}
    \begin{subfigure}[t]{0.48\linewidth}
      \vspace{0pt}
      \includegraphics[width=\textwidth]{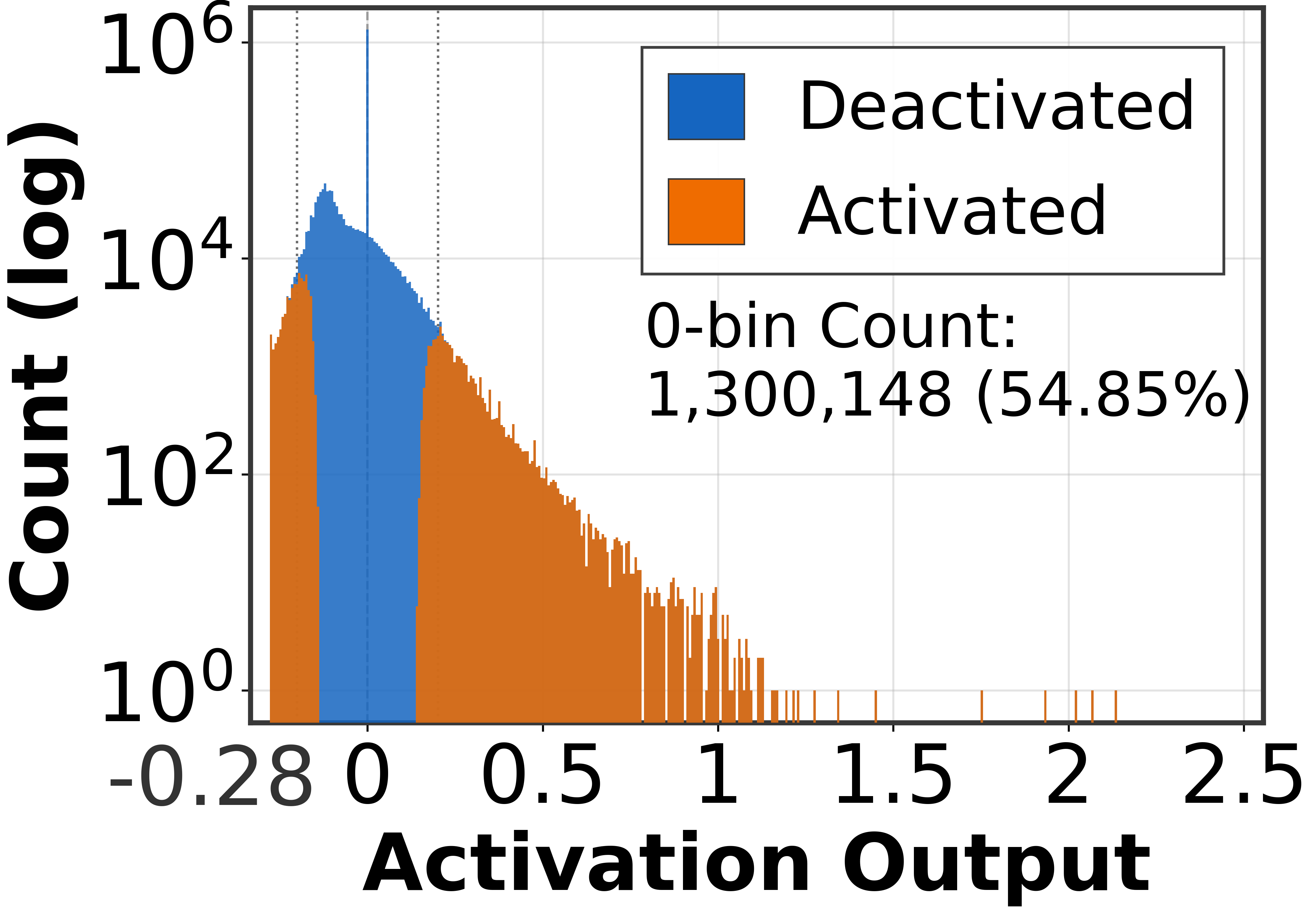}
      \caption{Activation Histogram}
      \label{fig:qwen_activation_histogram}
    \end{subfigure}\hfill
    \begin{subfigure}[t]{0.48\linewidth}
      \vspace{0pt}
      \includegraphics[width=\textwidth]{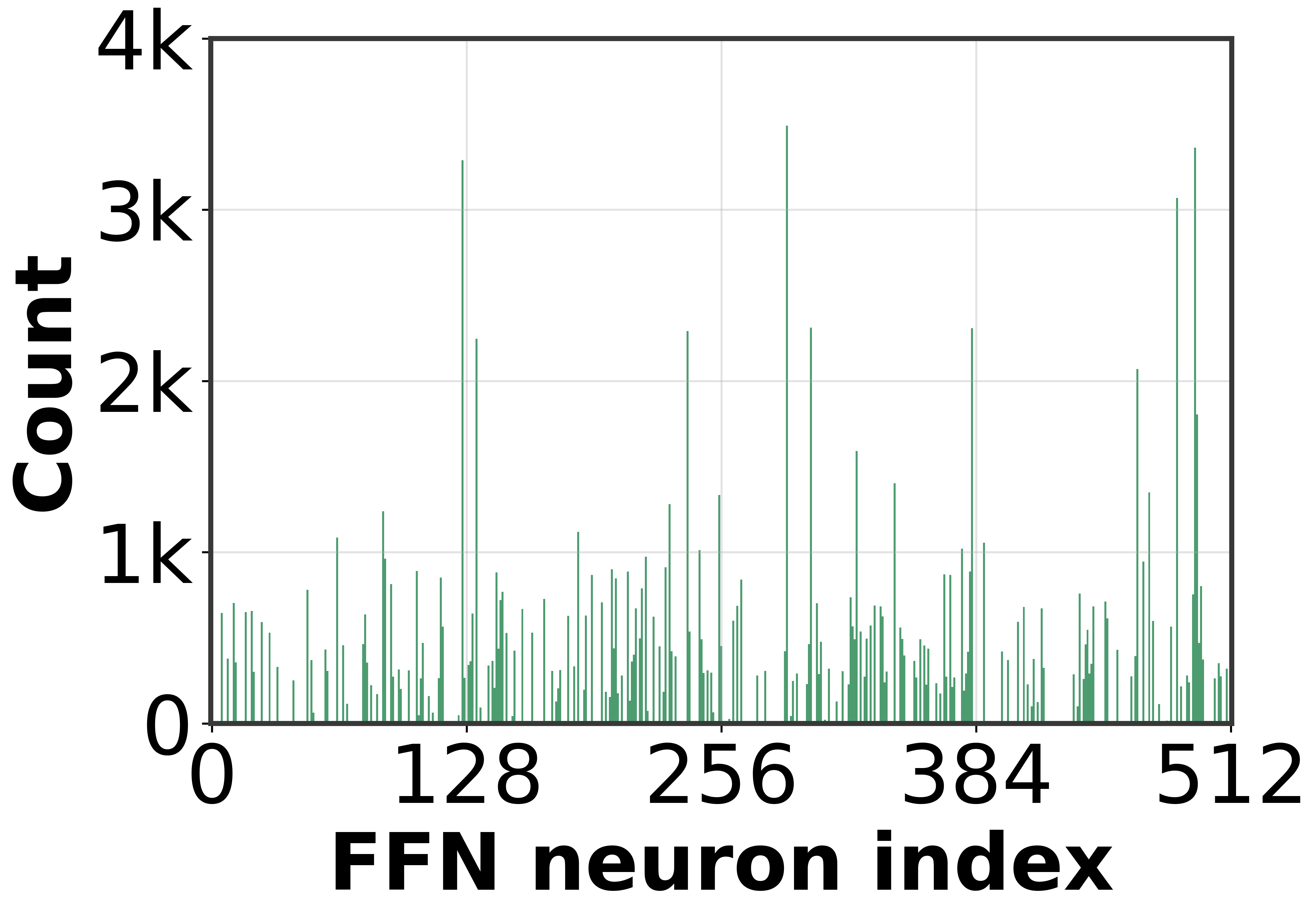}
      \caption{Per-neuron Count}
      \label{fig:qwen_per_neuron_count}
    \end{subfigure}
    \caption{Activation output histogram and per-neuron activation count of expert 0, layer 0 of Qwen3.5-35B-A3B at 95\% activation sparsity.}
    \label{fig:qwen_activation_details}
  \end{minipage}

  \vspace{-12pt}
\end{figure}

\noindent{\textbf{Neurons across Routed Experts:}}
To check if differences in neuron importance also exist within different routed experts, we distribute a \textit{neuron budget} across active routed experts. We sort and divide the activated experts into three groups $g_0,g_1,g_2$ based on their router weights (descending), with the remaining experts assigned to $g_2$ if the number of active experts $K$ is not divisible by three. Given a sparsity target $s$, we first calculate the total neuron budget $N = s\times K \times D$, where D is the FFN dimension of the model. Then, based on the distribution ratio $r_x$, we allocate $N_{g_x} = \tfrac{N\times r_x}{\Sigma{r_x\times |g_x|}}$ neurons to the experts of the group $g_x$, where $|g_x|$ is the number of experts in the group $g_x$.
Figure~\ref{fig:weighted_model_sparsity_accuracy} shows the resulting average accuracy under three different ratio configurations, shown in $r_0:r_1:r_2$ order. It is clear that, while allocating more neurons to experts with higher router priority does increase accuracy, \textbf{the benefits are minimal} with only up to 2 percentage points improvement.

This can be interpreted as a fine-grained analogue of expert skipping~\cite{lu2024notallexperts, zhong2024adapmoe}, which skips entire experts with low router scores for efficient MoE inference. While further studies on a more sophisticated neuron budgeting algorithm, such as router-weight-based waterfilling, may increase its benefits, we decide it is outside the scope of this paper. Therefore, we do not utilize neuron budgeting in our end-to-end system and leave it for future optimizations of intra-expert activation sparsity.

\noindent{\textbf{Neurons within a Single Expert:}}
To compare the impact of neurons within a single expert, we profile the SwiGLU activation output of expert 0, layer 0 of Qwen3.5-35B-A3B at 95\% sparsity on 2048 samples of the WikiText-2 dataset. Figure~\ref{fig:qwen_activation_histogram} presents the histogram of the activation outputs before and after masking the bottom 95\% activations, and Figure~\ref{fig:qwen_per_neuron_count} presents the activation count of each neuron after masking. Note that the y-axis of Figure~\ref{fig:qwen_activation_histogram} is on a logarithmic scale.

From Figure~\ref{fig:qwen_activation_histogram}, it is clear that the activation output of MoE expert FFN follows a long-tailed distribution, with an exponential decrease in count as the output magnitude increases. We also note the extreme spike in the 0-bin that includes the activations between -0.003 to 0.003, where \textbf{54.85\% of the total activation output is zero or near zero}. As such, even at 95\% sparsity, only the neurons with activation output between -0.2 to 0.2 are deactivated (blue), while the much larger range of activation output stays activated (orange). 

As seen from Figure~\ref{fig:qwen_per_neuron_count}, expert activation at 95\% sparsity is also extremely uneven across neurons with fewer than 10 neurons receiving more than 2000 activations, which is 8.5$\times$ the average of 235.12 activations. In addition, 284 neurons are never activated, roughly matching the 55\% observed in the 0-bin in Figure~\ref{fig:qwen_activation_histogram}. This observation shows that the neurons of MoE FFN layer are highly parameter inefficient and confirms the opportunity not only for intra-expert activation sparsity to reduce computation, but also for neuron pruning~\cite{lecun1989obd, ma2023llmpruner} to reduce model size as well. 

\section{Evaluation}
\label{sec:eval}

\subsection{Intra-expert Activation Sparse MoE in vLLM}
\label{sec:intra-expert-kernel}

Based on these observations, we integrate intra-expert activation sparsity in vLLM as a \textit{separate activation-sparse execution codepath} in the routed expert execution code, while leaving other parts of the system intact for compatibility and performance. The sparse execution path is activated when the user provides a \textbf{target activation sparsity} as an argument to the vLLM engine at launch. Full source code of our implementation is available in Appendix~\ref{apn:source_code} and supplementary materials.

\begin{figure}[t]
    \centering
      \begin{subfigure}[b]{0.172\textwidth}
    \includegraphics[width=\textwidth]{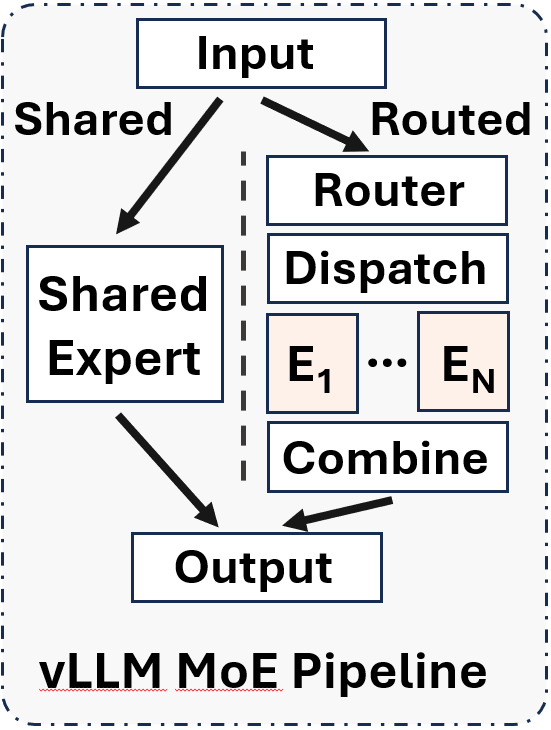}
    \caption{Moe Pipeline}
    \label{fig:moe_pipe}
  \end{subfigure}\hfill
    \begin{subfigure}[b]{0.205\textwidth}
    \includegraphics[width=\textwidth]{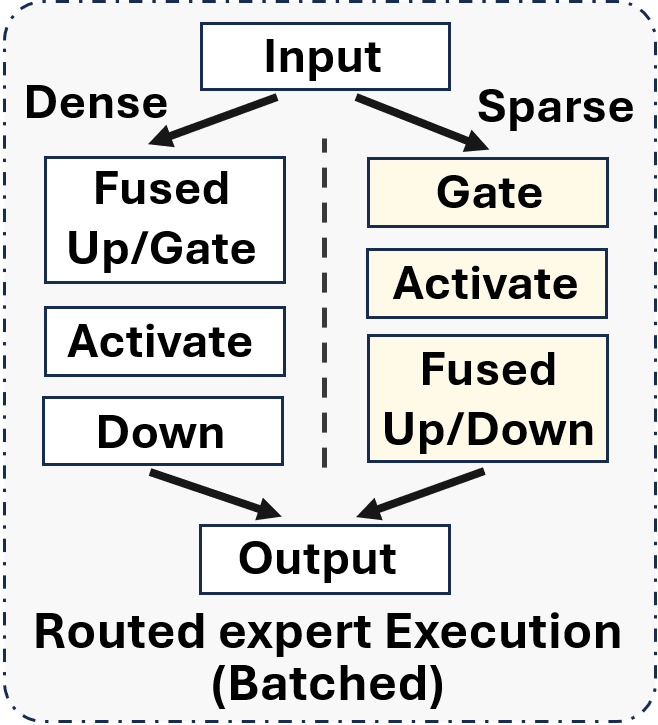}
    \caption{Sparse Codepath}
    \label{fig:per_expert_exec}
  \end{subfigure}\hfill
  \begin{subfigure}[b]{0.607\textwidth}
    \includegraphics[width=\textwidth]{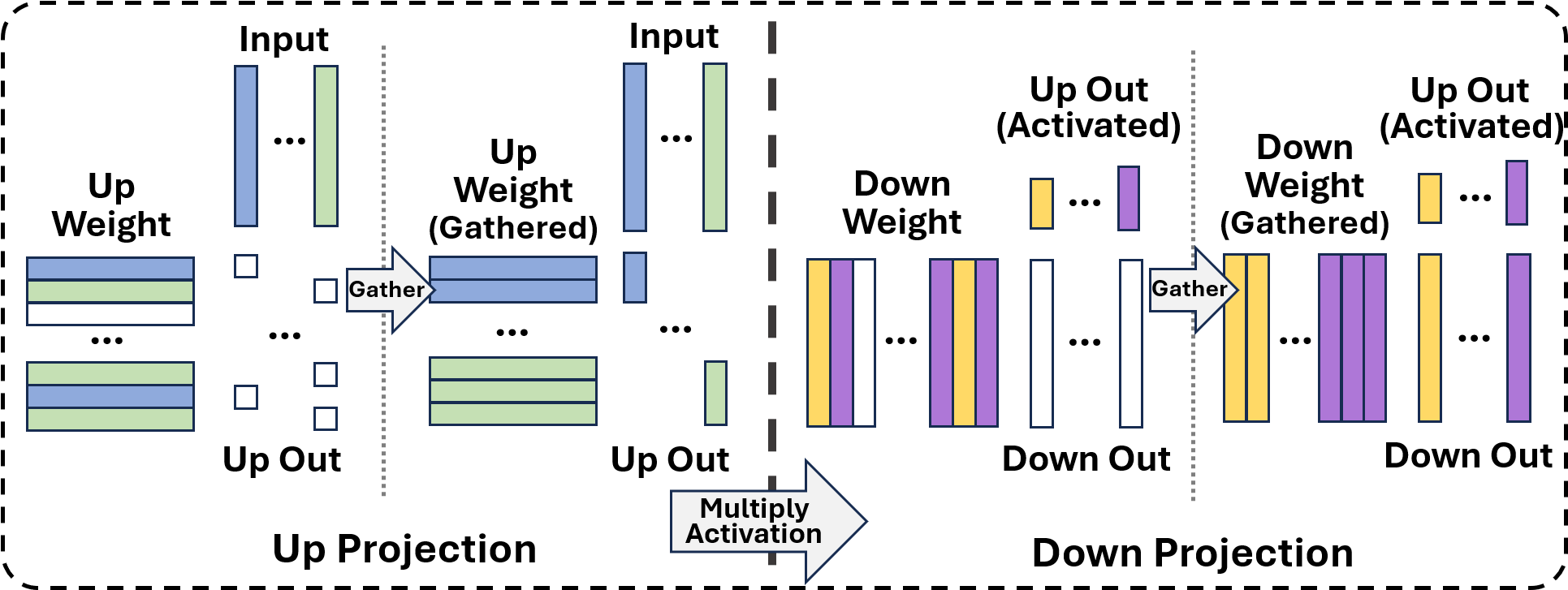}
    \caption{Fused Up-down Projection (Transposed for brevity)}
    \label{fig:fused_kernel}
  \end{subfigure}\hfill
    \caption{Overview of intra-expert activation sparsity integration in vLLM MoE execution pipeline.}
    \vspace{-20pt}
    \label{fig:kernel_design}
\end{figure}

As shown in the vLLM MoE Pipeline of Figure~\ref{fig:moe_pipe}, we exclusively modify the routed per-expert execution code highlighted in orange without changing shared expert execution or the router, dispatch, and combine functions that surround the per-expert execution. Applying activation sparsity to the shared expert has little to no gain, as they are executed in parallel to the routed expert path in the vLLM engine, and their neurons have a high impact on accuracy. Similarly, router, dispatch, and combine functions in routed execution are critical to the performance of expert parallelism (EP) and often use hand-crafted all-to-all communication libraries such as RCCL~\cite{amd2020rccl}, NCCL~\cite{nvidia2015nccl}, or DeepEP~\cite{zhao2025deepep}, which cannot be easily modified without severely impacting parallelism performance. 

Figure~\ref{fig:per_expert_exec} visualizes the changes we made to the routed expert execution. We add the activation sparse execution path on the right, highlighted in yellow. The execution path can be dynamically switched between the original dense expert execution of vLLM and our sparse execution path, depending on the step batch size, as explained later. Note that the operators such as up, gate, activate, and down projections are batched across the entire routed experts $E_1$ to $E_N$ in the given EP rank for efficiency in both dense vLLM and our sparse execution path, which is skipped in the figure for brevity.
Our sparse code has \textbf{three key characteristics} that differ from the dense execution: (1) Standalone gate projection, (2) Custom sparse activation kernel, and (3) Fused sparse up-down projection kernel. 

Gate projection must be executed before all sparse operations, as its output is necessary to compute the neuron activations and sparsity mask. To efficiently compute gate projection, we reuse the \textit{batched matrix multiplication} (BMM) kernel used for the down projection of the dense path, which provides high-throughput expert-major computation and can be easily repurposed by providing the gate weights as input. As the gate projection uses a dense BMM kernel, it limits the scope of sparse computation to roughly two-thirds of the execution (up and down), and maximum speedup to 3$\times$. However, we find that overall execution still yields substantial performance improvements, as shown in Section~\ref{sec:eval_kernel}. Nevertheless, further studies on fast approximation or activation sparsity mask reuse, similar to those in sparse KV cache~\cite{zhang2023h2o, li2024snapkv}, may allow for a fully sparse expert execution and enable higher performance in intra-expert activation sparsity.

We implement a custom sparse activation kernel that fuses the activation calculation and masking of the deactivated neurons into a single kernel using Triton~\cite{triton_program_language_api}. The kernel computes neuron activations using the gate projection output and applies threshold-based single-pass masking~\cite{liu2024teal, lee2024cats}. The user-supplied sparsity argument is turned into a model-specific threshold using a lookup table generated during engine startup. If a neuron's absolute activation output value is below this threshold, the neuron is masked out. This allows for an $\mathcal{O}(n)$ sparsity mask generation compared to the traditional $\mathcal{O}(n \log n)$ top-k approach, with minimal accuracy loss as shown in Section~\ref{sec:eval_breakdown}. The activation kernel then gathers the activated neuron indices using the mask and returns it in a continuous buffer. 

Finally, we create a fused sparse up-down projection kernel based on the Triton BMM kernel provided by vLLM. As shown in Figure~\ref{fig:fused_kernel}, we first use batched vector load to efficiently \textit{gather the up weight vectors} of each activated neuron for a given input token, using the active neuron index buffer generated by the activation kernel. This creates a packed, dense up weight matrix on the shared memory of the GPU, which can be efficiently computed using a dense matrix-vector multiplication against the input token to generate the up projection output. The output is multiplied by the activation and used as an input to the matrix-vector multiplication with the down-projection weights, which are also gathered and packed into a dense matrix using batched vector loads. 

To maintain compatibility with Hip Graph and CUDA Graph, which require reuse of the same GPU memory buffers and kernel launch parameters across graph invocations, the gathered active neuron index buffer is padded into the full neuron size, and the Triton kernels are launched with a tile granularity of a fixed 64 neurons per Triton program. Each Triton program first checks its assigned neuron tile from the active neuron index vector, and exits early if its assigned tile only contains padding indices. This allows for efficient implementation of sparse kernel execution while keeping compatibility with graph capture optimizations, and the fixed 64 neuron tile granularity enables utilization of GPU hardware accelerators for batched memory access and high-throughput computation instructions. 

This approach has a limited maximum throughput compared to the dense BMM in the compute-bound region with large batch sizes. This is a fundamental limitation of a gather-based sparse execution solution, as GPU hardware provides the highest throughput when executing dense operations with contiguous memory access. To overcome this limitation, we profile the tipping batch size between the sparse execution path and dense execution path during engine startup. Based on this, we switch between sparse and dense execution paths based on the batch size of the running step. This allows us to use the low-latency sparse path when there are fewer workloads, and switch back to the high-throughput dense path when there are more workloads.

\subsection{Performance Evaluation}

\subsubsection{MoE Layer Performance}
\label{sec:eval_kernel}

\begin{figure*}[t]
    \centering
    \hfill
    \begin{subfigure}[c]{0.32\textwidth}
        \caption{Qwen3.5-35B-A3B, MI355X}
        \includegraphics[width=\linewidth]{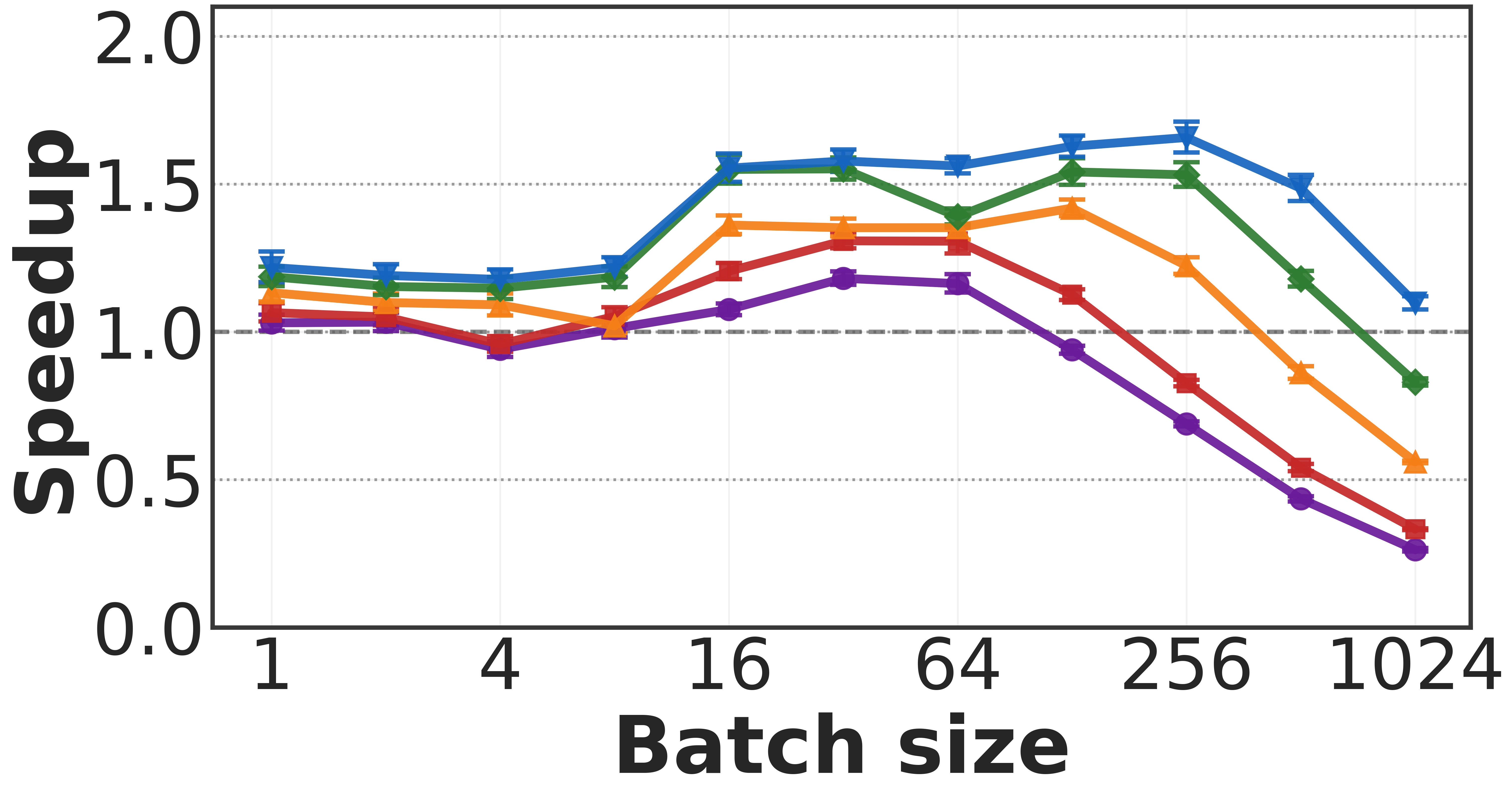}
    \end{subfigure}\hfill
    \begin{subfigure}[c]{0.32\textwidth}
    \caption{Qwen3.5-35B-A3B, H200}
        \includegraphics[width=\linewidth]{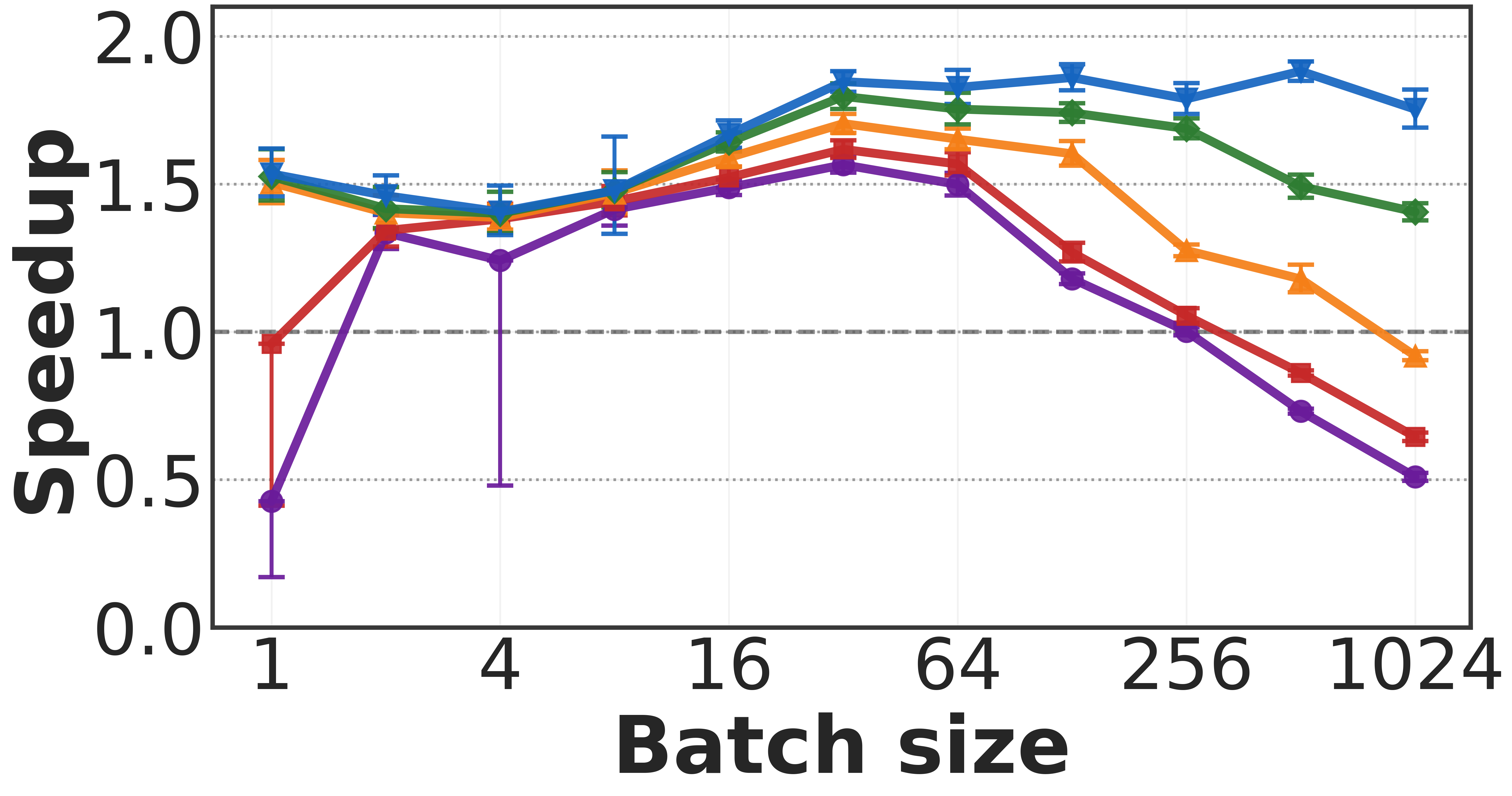}
    \end{subfigure}\hfill
    \begin{subfigure}[c]{0.32\textwidth}
        \caption{Qwen3.5-35B-A3B, RTX4090}
        \includegraphics[width=\linewidth]{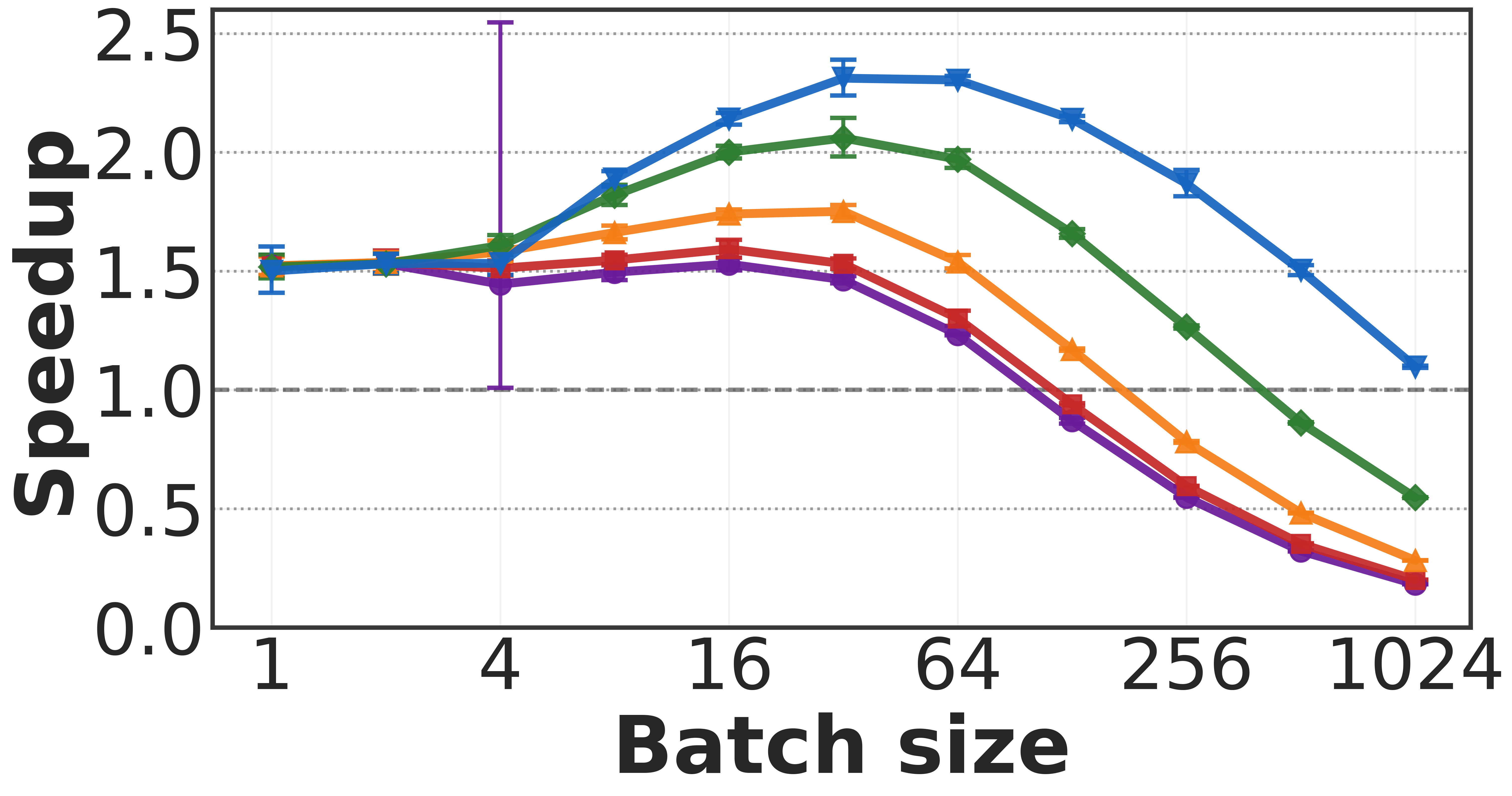}     
    \end{subfigure}\hfill
    
    \hfill
    \begin{subfigure}[c]{0.32\textwidth}
        \caption{Qwen3.5-122B-A10B, MI355X}
        \includegraphics[width=\linewidth]{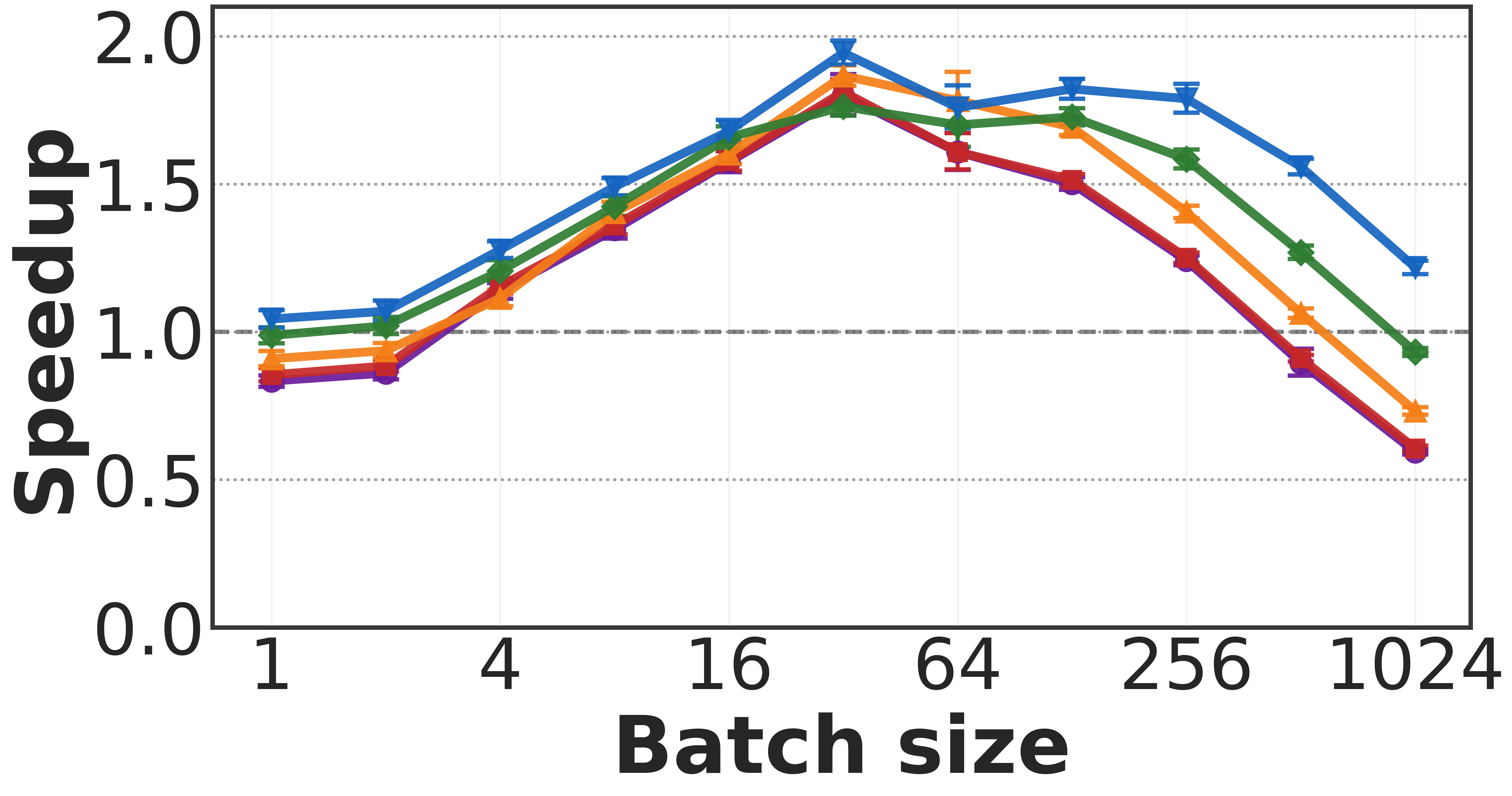}
    \end{subfigure}\hfill
    \begin{subfigure}[c]{0.32\textwidth}
        \caption{Qwen3.5-122B-A10B, H200}
        \includegraphics[width=\linewidth]{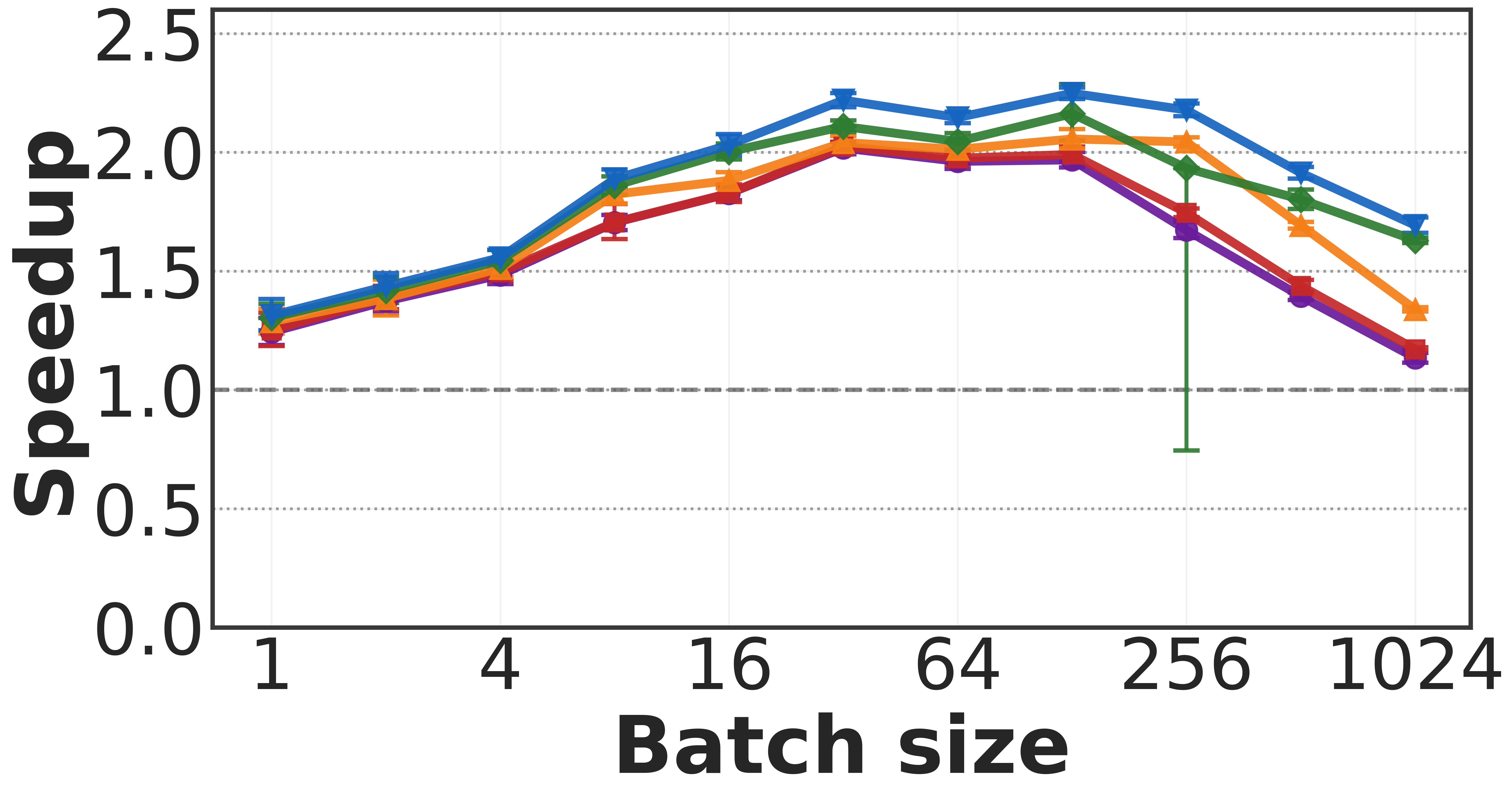}
    \end{subfigure}\hfill
    \begin{subfigure}[c]{0.32\textwidth}
        \caption{Qwen3.5-122B-A10B, RTX4090}
        \includegraphics[width=\linewidth]{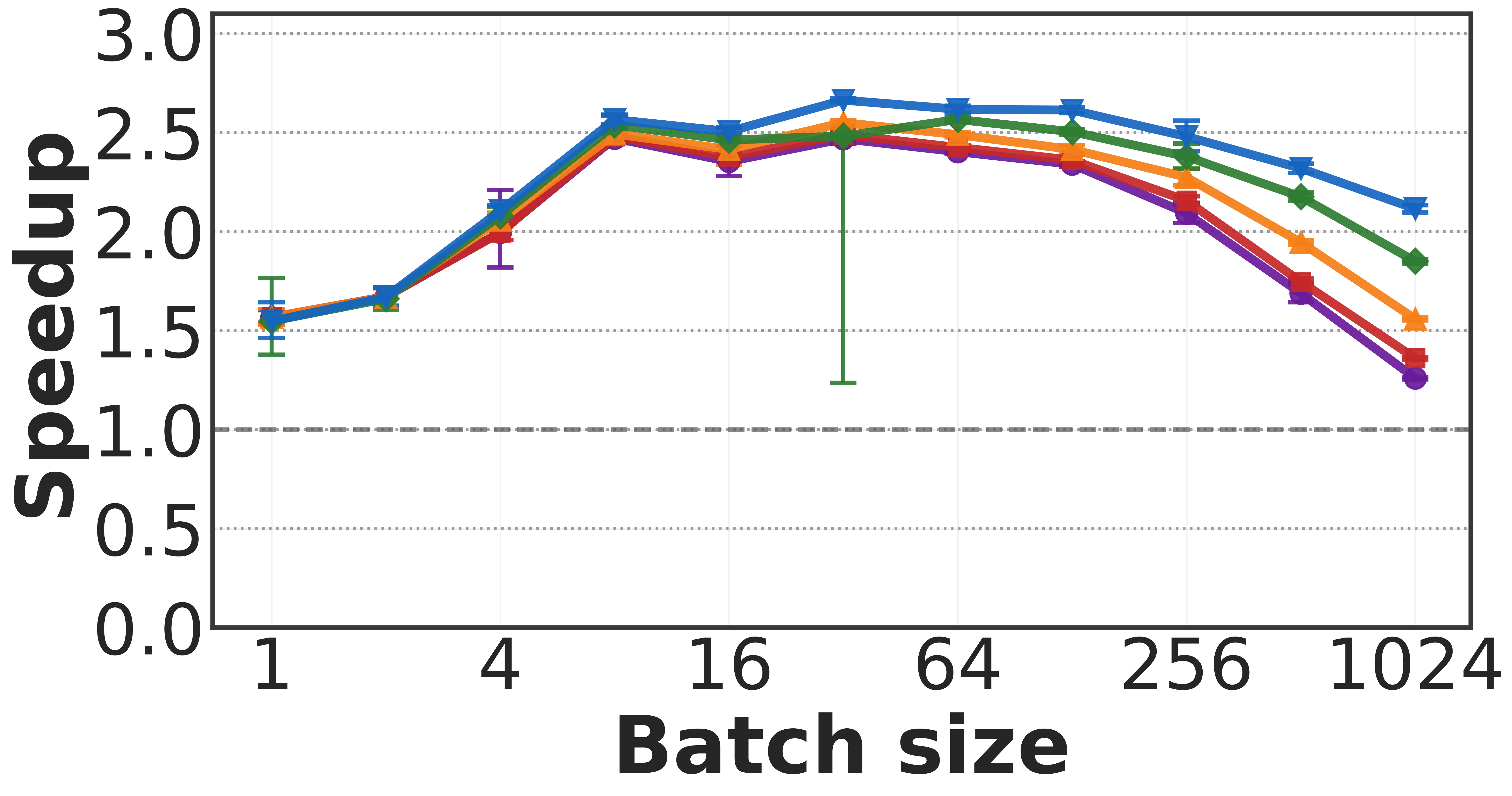}
    \end{subfigure}\hfill

    \includegraphics[width=0.75\textwidth]{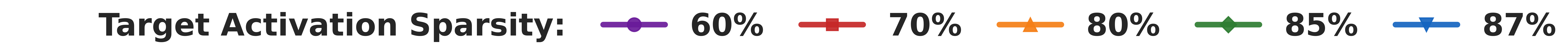}

    \caption{MoE layer speedup of intra-expert activation sparse execution against dense vLLM on various settings, averaged across 100 runs with a 96\% confidence interval. Note that Qwen3.5-35B-A3B and Qwen3.5-122B-A10B have a sparsity cutoff of \textbf{84.5\% and 87.4\%}, respectively.}
    \label{fig:speedup}
    \vspace{-12pt}
\end{figure*}

We integrate our intra-expert activation sparse MoE execution on a development build of vLLM~\cite{kwon2023efficientmemorymanagementlarge}\footnote{Version 0.1.dev15965+g6506aec54}, and evaluate its benefits by comparing its speedup against the original dense vLLM baseline. We use AMD MI355X, NVIDIA H200, and NVIDIA RTX4090 GPUs for our experiments. 
We use Qwen3.5-35B-A3B and Qwen3.5-122B-A10B models~\cite{yang2025qwen3technicalreport}, with samples from WikiText-2~\cite{merity2016wikitext} dataset as input to our experiments. We use the total activation sparsity levels of 60\%, 70\%, 80\%, 85\%, and 87\%, covering a representative set of the model sparsity cutoff values observed in Section~\ref{sec:intra-expert}. 

Figure~\ref{fig:speedup} presents the speedup of intra-expert activation sparse MoE execution over the dense vLLM baseline, measured while executing the first MoE layer of the given model. The sparse execution achieves up to 1.5--2.5$\times$ speedup at the 85\% and 87\% sparsity cutoff for Qwen3.5-35B-A3B and Qwen3.5-122B-A10B. As expected, the speedup is highly dependent on batch size, with peak speedup observed between batch sizes of 16 to 128. At smaller batch sizes, static overhead such as kernel launch cost dominates, offsetting the computation savings of sparse execution and giving flat speedup. At larger batch sizes, execution approaches the compute-bound regime, where the higher peak throughput of the dense execution outweighs the benefits of sparsity. Together, these two effects cause the speedup curve to form a convex shape across batch sizes. 

Sparsity plays a key role in the speedup, as both the peak speedup and the range of batch sizes with speedup scale with the degree of sparsity.
To be specific, at smaller batch sizes the speedup differences across sparsity levels are less pronounced, as the absolute reduction in computation is small and static overheads such as the dense gate projection and activation kernel dominate. As batch size increases and execution approaches the compute-bound regime, the degree of sparsity plays a more decisive role: lower sparsity levels of 60--70\% see a sharper decline in speedup, while higher sparsity levels of 85--87\% remain more robust across a wider batch size range.

Hardware and model size also greatly affect the efficacy of sparse execution. The general trend is that \textbf{sparse execution provides greater speedup when resource bottlenecks are more severe}. Comparing the MI355X, H200, and RTX4090---where the MI355X has the most raw compute and memory resources and the RTX4090 has the least---we find that the RTX4090 achieves the highest speedup of up to 2.5$\times$, while the MI355X reaches only up to 1.8$\times$, with the H200 in between at 2.0$\times$. Similarly, we observe roughly 30--40 percentage points increases in speedup across all configurations on Qwen3.5-122B-A10B compared to Qwen3.5-35B-A3B, confirming the greater benefits of sparse execution on more compute- and memory-intensive workloads. As such, the highest speedup of 2.5$\times$ is observed while running the larger Qwen3.5-122B-A10B MoE layer on the resource-constrained RTX4090 GPU.

\subsubsection{Sparsity, Accuracy, and Execution Breakdown}
\label{sec:eval_breakdown}

\begin{table*}[t]
  \centering
  \begin{minipage}[t]{0.63\textwidth}
    \vspace{0pt} 
    \centering
    \resizebox{\linewidth}{!}{
    \begin{tabular}{cccccc}
      \toprule
      \multicolumn{1}{c}{\begin{tabular}[c]{@{}c@{}}Target Actv.\\
      Sparsity\end{tabular}} & 
      \multicolumn{1}{c}{\begin{tabular}[c]{@{}c@{}}Total
      Actv.\\Sparsity\end{tabular}} & 
      \multicolumn{1}{c}{\begin{tabular}[c]{@{}c@{}}Routed
      Expt.\\Sparsity\end{tabular}} & 
      \multicolumn{1}{c}{\begin{tabular}[c]{@{}c@{}}Average\\
      Accuracy\end{tabular}} & 
      \multicolumn{1}{c}{\begin{tabular}[c]{@{}c@{}}Time\\
      Taken\end{tabular}} & 
      \multicolumn{1}{c}{Speedup} \\
      \midrule
      \textbf{Dense} & - & - & 70.9\% & 0.354ms & - \\
      \textbf{60\%} & 57\% & 64\% & 68.2\% & 0.383ms & 0.92$\times$ \\
      \textbf{70\%} & 72\% & 81\% & 67.6\% & 0.315ms & 1.12$\times$ \\
      \textbf{80\%} & 79\% & 89\% & 67.7\% & 0.249ms & 1.42$\times$ \\
      \textbf{85\%} & 84\% & 94\% & 67.5\% & 0.229ms & 1.55$\times$ \\
      \textbf{87\%} & 87\% & 98\% & 65.7\% & 0.197ms & 1.8$\times$ \\
      \bottomrule
    \end{tabular}
    }
    \caption{Qwen3.5-35B-A3B achieved total activation and routed expert sparsity, average 
    benchmark accuracy, and time taken on various target activation sparsity on MI355X, Batch size=128}
    \label{tab:target_sparsity_performance}
  \end{minipage}\hfill
  \begin{minipage}[t]{0.35\textwidth}
    \vspace{2pt} 
    \centering
    \includegraphics[width=\linewidth]{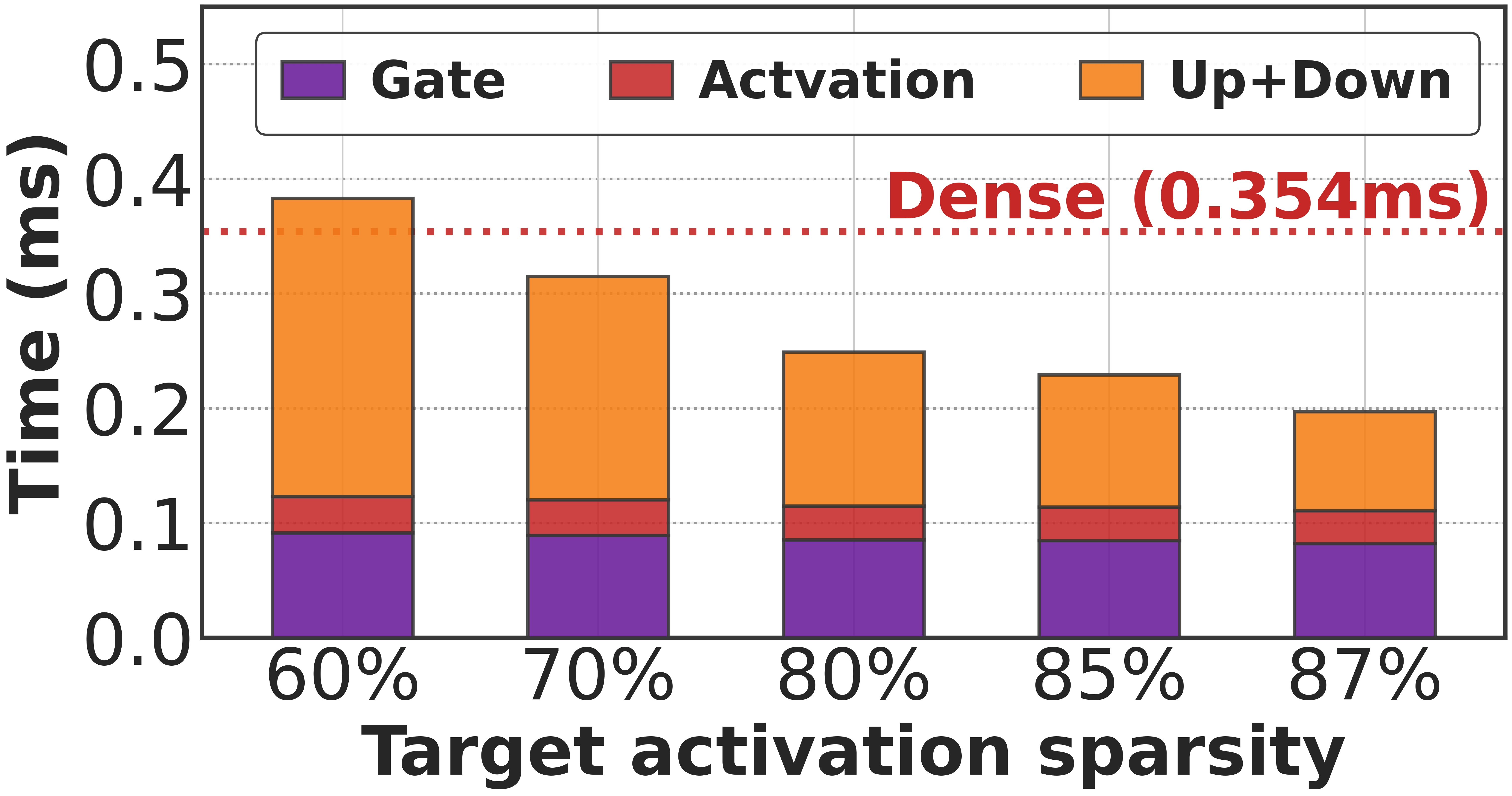}
    \captionof{figure}{Qwen3.5-35B-A3B MoE layer sparse execution time breakdown on MI355X, Batch size=128}
    \label{fig:time_breakdown}
  \end{minipage}
  \vspace{-12pt}
\end{table*}

Table~\ref{tab:target_sparsity_performance} presents the achieved sparsity and end-to-end average benchmark accuracy of our threshold-based activation sparse MoE execution, running the same set of benchmarks as in Section~\ref{sec:intra-expert-analysis}. As described in Section~\ref{sec:intra-expert-kernel}, we use a lookup table to convert the user-specified target activation sparsity to a per-neuron deactivation threshold. This threshold-based masking is applied to routed experts, yielding a model-specific routed expert sparsity that translates to the desired total activation sparsity when shared experts are accounted for.

As shown in the Total Activ. Sparsity column of the table, our threshold-based approach achieves total activation sparsity within $\pm3$ percentage points of the desired target sparsity. Furthermore, the Average Accuracy column shows that the sparse execution maintains 95\% of the baseline average benchmark score up to the target sparsity of 85\%. This closely matches the model's sparsity cutoff identified in our earlier analysis and confirms that \textbf{our sparse execution correctly reproduces the activation sparse model accuracy} observed in Section~\ref{sec:intra-expert-analysis}.

Figure~\ref{fig:time_breakdown} shows the time breakdown of the sparse MoE execution across different target activation sparsity levels. We see that the dense gate projection and activation kernel contribute a constant overhead to execution, and only the gather-based fused up-down projection kernel benefits from increased sparsity. As noted previously, this is a limitation of our current approach: earlier-stage sparsity map generation or reuse would allow sparsity to be applied to the gate projection as well, enabling further performance gains. Nevertheless, our threshold-based sparse execution still achieves 1.55$\times$ speedup at the model sparsity cutoff, as shown in Table~\ref{tab:target_sparsity_performance}, demonstrating substantial benefits to inference performance.

\subsection{End-to-end Evaluation}
\label{sec:eval_e2e}

Table~\ref{tab:e2e_speedup} presents the end-to-end model inference speedup of intra-expert activation sparse MoE execution over the dense vLLM baseline. Note that batch-size-based switching between dense and sparse MoE execution is enabled in the evaluation, guaranteeing speedup across all configurations. With the switch enabled, we observe up to 1.2$\times$ speedup on H200 and up to 1.1$\times$ on MI355X, with a minimum speedup of 1.0$\times$ as expected. Sparse execution primarily benefits latency-bound workloads, such as decode-heavy configurations with large numbers of output tokens or small numbers of requests. In compute-bound scenarios such as prefill-heavy workloads or executions with large numbers of requests, the benefits are reduced, as the per-step batch size frequently exceeds the switching threshold and the dense path is utilized more often.

\begin{table*}[t]
  \centering
  \begin{minipage}[t]{0.73\textwidth}
    \vspace{0pt} 
    \centering
    \resizebox{\linewidth}{!}{
    \begin{tabular}{cccccccccc}
      \toprule                                                                                                 
      \multirow{2}{*}{\begin{tabular}[c]{@{}c@{}}GPU\end{tabular}} &
      \multirow{2}{*}{\begin{tabular}[c]{@{}c@{}}Input\\Tokens\end{tabular}} &
      \multirow{2}{*}{\begin{tabular}[c]{@{}c@{}}Output\\Tokens\end{tabular}} &
      \multicolumn{6}{c}{Number of Requests} &
      \multirow{2}{*}{\begin{tabular}[c]{@{}c@{}}Average\\Speedup\end{tabular}} \\
      \cmidrule(lr){4-9}
      & & & 1 & 4 & 16 & 64 & 256 & 512 & \\
      \midrule                          
       \multirow{5}{*}{MI355X}
       & 1024 & 1    & $1.01\times$ & $1.01\times$ & $1.01\times$ & $1.01\times$ & $1.00\times$ & $1.00\times$ & $1.01\times$ \\
       & 1024 & 128  & $1.10\times$ & $1.08\times$ & $1.10\times$ & $1.06\times$ & $1.03\times$ & $1.00\times$ & $1.06\times$ \\
       & 1    & 1024 & $1.04\times$ & $1.03\times$ & $1.05\times$ & $1.06\times$ & $1.07\times$ & $1.00\times$ & $1.04\times$ \\
       & 128  & 1024 & $1.07\times$ & $1.07\times$ & $1.08\times$ & $1.07\times$ & $1.10\times$ & $1.00\times$ & $1.08\times$ \\
       & 1024 & 1024 & $1.09\times$ & $1.09\times$ & $1.09\times$ & $1.07\times$ & $1.05\times$ & $1.00\times$ & $1.07\times$ \\
      \midrule                                                        
      \multirow{5}{*}{H200}
       & 1024 & 1    & $1.11\times$ & $1.09\times$ & $1.08\times$ & $1.09\times$ & $1.08\times$ & $1.10\times$ & $1.09\times$ \\
       & 1024 & 128  & $1.17\times$ & $1.15\times$ & $1.13\times$ & $1.08\times$ & $1.06\times$ & $1.06\times$ & $1.11\times$ \\
       & 1    & 1024 & $1.18\times$ & $1.19\times$ & $1.19\times$ & $1.19\times$ & $1.19\times$ & $1.13\times$ & $1.18\times$ \\
       & 128  & 1024 & $1.19\times$ & $1.18\times$ & $1.19\times$ & $1.19\times$ & $1.17\times$ & $1.13\times$ & $1.18\times$ \\
       & 1024 & 1024 & $1.20\times$ & $1.20\times$ & $1.20\times$ & $1.19\times$ & $1.14\times$ & $1.13\times$ & $1.18\times$ \\
      \bottomrule                                                     
      \end{tabular}
      }
      \caption{End-to-end Qwen3.5-35B-A3B inference speedup of activation sparse MoE execution at 85\% target sparsity against dense vLLM baseline.}     
    \label{tab:e2e_speedup}
  \end{minipage}\hfill
  \begin{minipage}[t]{0.245\textwidth}
    \vspace{0pt} 
    \centering
    \includegraphics[width=\linewidth]{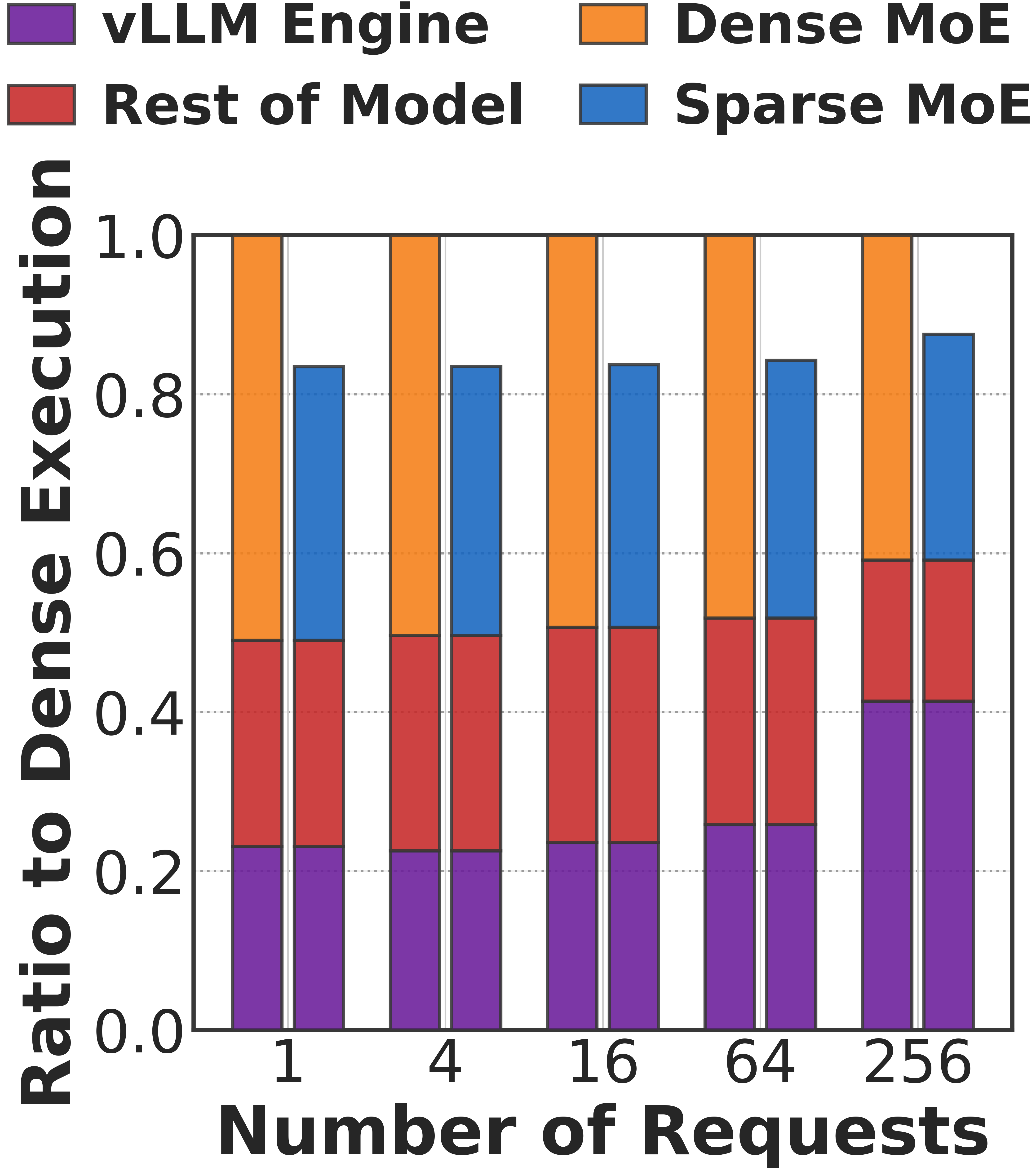}
    \captionof{figure}{End-to-end execution time ratio.}
    \label{fig:e2e_breakdown}
  \end{minipage}
  \vspace{-12pt}
\end{table*}

Figure~\ref{fig:e2e_breakdown} shows the execution time ratio of each end-to-end operation compared to the full dense vLLM execution, on H200 for end-to-end Qwen3.5-35B-A3B inference on requests with 1024 input and output tokens. ``Dense MoE'' and ``Sparse MoE'' refer to routed expert execution time measured in Section~\ref{sec:eval_kernel}. ``Rest of Model'' is the remaining model components such as the attention or embedding layers. ``vLLM Engine'' refers to the serving overhead of vLLM, including request scheduling, KV cache management, and output sampling.

Across batch sizes, dense MoE layer execution time accounts for roughly 45\% of total execution time. While we achieve a 1.5--1.8$\times$ speedup in the MoE layer, Amdahl's law~\cite{amdahl1967validity} limits the end-to-end improvement to roughly a 15--20\% reduction in total execution time, which corresponds to the 1.2$\times$ end-to-end speedup observed in Table~\ref{tab:e2e_speedup}. Nevertheless, the MoE layer constitutes the majority share of model execution time across all batch sizes, making it a high-impact target for optimization.

\section{Discussion}
\label{sec:discussion}

\noindent{\textbf{Sparse Gate Projection: }} 
The current design applies activation sparsity only on the up and down projections, which account for roughly two-thirds of the expert FFN computation. This imposes a theoretical upper bound of 3$\times$ on end-to-end speedup even under optimal up/down execution. Thus, future work could explore ways to allow a lightweight gate-activation prediction mechanism that estimates sparse activation patterns before computing the full gate projection, or reusing the sparsity pattern based on the input token. Achieving accurate prediction of the sparsity map before gate projection would allow the system to prevent full dense gate computation, reducing activation overhead and enabling performance gains closer to the theoretical upper bound of intra-expert sparsity, enabling higher end-to-end MoE speedups.

\noindent{\textbf{Grouped Intra-expert Activation Sparsity: }} 
Our results reveal that sparsity alone does not guarantee speedup, as mapping sparse computation efficiently onto GPU hardware introduces overhead due to irregular memory access, indexing, and limited hardware utilization. This gap between theoretical sparsity and realized speedup widens as batch size increases, underscoring the need for more optimized sparse kernel designs. The most promising direction is to increase the granularity of sparsity mapping: rather than assigning and gathering each individual active neuron, tokens and neurons can be grouped to fit the hardware accelerator width and perform sparsity mapping at this block level. Minimal group sizes between 4 and 32 would better align sparse computation with the regular tile structures expected by high-performance kernels, with only a minor tradeoff in precision, leading to improved memory coalescing and increased hardware utilization.

\section{Conclusion}
In this work, we demonstrated how intra-expert activation sparsity can improve the efficiency of MoE models. While conventional MoE architectures reduce computation by limiting the number of activated experts per token, our approach further exploits sparsity in the dense FFN of each selected expert. We identified substantial redundancy within the dense expert itself and demonstrated that MoE inference can be further accelerated at inference time. Intra-expert activation sparsity is a promising and practical complement to inter-expert sparsity for reducing the serving cost of large MoE models, with growing relevance as model scale and inference demand continue to rise.

\label{sec:conclusion}


\bibliographystyle{plain}
\bibliography{references}

\appendix
\newpage
\appendix
\onecolumn
\section{Licenses of the assets used}
\label{apn:licenses}
\begin{table}[h]
\centering
\caption{Licenses of the assets used in the paper.}
\label{tab:licenses}
\resizebox{\textwidth}{!}{
\begin{tabular}{llll}
\toprule
\textbf{Asset} & \textbf{Type} & \textbf{License} & \textbf{URL} \\
\midrule
Granite-1B-A400M    & Model   & Apache 2.0                  & \url{https://huggingface.co/ibm-granite/granite-3.0-1b-a400m-base} \\
OLMoE-1B-7B         & Model   & Apache 2.0                  & \url{https://huggingface.co/allenai/OLMoE-1B-7B-0924} \\
DeepSeek-V2-Lite    & Model   & DeepSeek Model License      & \url{https://huggingface.co/deepseek-ai/DeepSeek-V2-Lite} \\
GPT-OSS-20B         & Model   & Apache 2.0                  & \url{https://huggingface.co/openai/gpt-oss-20b} \\
Qwen3.5-35B-A3B     & Model   & Apache 2.0                  & \url{https://huggingface.co/Qwen/Qwen3.5-35B-A3B} \\
Qwen3.5-122B-A10B   & Model   & Apache 2.0                  & \url{https://huggingface.co/Qwen/Qwen3.5-122B-A10B} \\
Qwen3.5-397B-A17B   & Model   & Apache 2.0                  & \url{https://huggingface.co/Qwen/Qwen3.5-397B-A17B} \\
Llama-4-Maverick    & Model   & Llama 4 Community License   & \url{https://huggingface.co/meta-llama/Llama-4-Maverick-17B-128E} \\
\midrule
ARC-Challenge       & Benchmark & Apache 2.0               & \url{https://huggingface.co/datasets/allenai/ai2_arc} \\
ARC-Easy            & Benchmark & Apache 2.0               & \url{https://huggingface.co/datasets/allenai/ai2_arc} \\
HellaSwag           & Benchmark & MIT                      & \url{https://huggingface.co/datasets/Rowan/hellaswag} \\
Winogrande          & Benchmark & Apache 2.0               & \url{https://huggingface.co/datasets/allenai/winogrande} \\
TruthfulQA          & Benchmark & Apache 2.0               & \url{https://huggingface.co/datasets/truthful_qa} \\
WikiText-2          & Dataset   & CC BY-SA 4.0             & \url{https://huggingface.co/datasets/Salesforce/wikitext} \\
\midrule
vLLM                & Framework & Apache 2.0               & \url{https://github.com/vllm-project/vllm/blob/main/LICENSE} \\
lm-evaluation-harness & Framework & MIT                   & \url{https://github.com/EleutherAI/lm-evaluation-harness/blob/main/LICENSE} \\
\bottomrule
\end{tabular}
}
\end{table}

\section{Intra-expert Activation Sparse MoE Execution Source Code}
\label{apn:source_code}

\begin{lstlisting}[language=Python, caption=Full source code of sparse MoE execution kernel.]
"""Gate-sparse MoE kernel integration for vLLM.

Import inside vllm/mode_executor/layers/fused_moe/layer.py and inject 
sparse_moe_forward() at the forward_native() function of the FusedMoE class 
in layer.py for sparse execution.

Router computation (softmax + top-k) is performed by vLLM standard router
(DefaultMoERunner.router.select_experts) before this module is called;
topk_weights and topk_ids are passed in directly.

"""

from __future__ import annotations

import functools
import logging
import os
import torch
import triton

from vllm import _custom_ops as ops
from vllm.model_executor.layers.fused_moe.fused_moe import (
    invoke_fused_moe_triton_kernel,
    try_get_optimal_moe_config,
    _get_config_dtype_str,
)

logger = logging.getLogger(__name__)

# ---------------------------------------------------------------------------
# One-shot sparsity diagnostic
# ---------------------------------------------------------------------------
_printed_sparse_map: set = set()


def _maybe_print_sparse_map(
    layer_id: int,
    sigma: float,
    gate_raw_flat: "torch.Tensor",  # [T*K, N]  post-gate-GEMM, pre-threshold
    T: int,
    K: int,
    N: int,
) -> None:
    """Print mask[0] for token-0 expert-0 once per (layer_id, gate_cutoff) pair."""
    key = (layer_id, round(sigma, 6))
    if key in _printed_sparse_map:
        return
    g = gate_raw_flat[0].float().cpu()          # token-0, expert-0, shape [N]
    mask = (g >= sigma).to(torch.int8)
    n_active = int(mask.sum().item())
    # Heuristic: skip one-shot when every channel passes a strictly positive cutoff
    # (often uninteresting for diagnostics; old z-score path also used this for warmup).
    if n_active == N and sigma > 0.0:
        return
    _printed_sparse_map.add(key)
    frac = n_active / N
    print(
        f"[sparse_moe] layer={layer_id} sigma={sigma:.2f}  "
        f"mask[0]: {n_active}/{N} active  frac={frac:.4f}  "
        f"{mask.tolist()}",
        flush=True,
    )

_CAPTURE_LAYER_ID: int | None = (
    int(os.environ["SPARSE_MOE_CAPTURE_LAYER"])
    if "SPARSE_MOE_CAPTURE_LAYER" in os.environ else None
)
_CAPTURE_T: int = int(os.environ.get("SPARSE_MOE_CAPTURE_T", "32"))
_CAPTURE_PATH: str = os.environ.get("SPARSE_MOE_CAPTURE_PATH", "")
_capture_done: bool = False

try:
    import triton.language as tl
    _SPARSE_MOE_AVAILABLE = True
except Exception as _e:
    _SPARSE_MOE_AVAILABLE = False
    logger.warning(
        "Gate-sparse MoE kernel not available: %s. "
        "Falling back to dense FusedMoE.", _e,
    )

# Maximum token count for which the sparse kernel is engaged.
# Must match _SPARSE_MAX_T in layer.py.
_SPARSE_MAX_T: int = 384

# ---------------------------------------------------------------------------
# Utility functions
# ---------------------------------------------------------------------------

def _next_pow2(n: int) -> int:
    return 1 << (n - 1).bit_length()


def _estimate_k_total(N: int, K: int, block_k: int = 32) -> int:
    k = K * N
    return ((k + block_k - 1) // block_k) * block_k


_SM_COUNT: int | None = None


def _sm_count() -> int:
    global _SM_COUNT
    if _SM_COUNT is None:
        _SM_COUNT = torch.cuda.get_device_properties(
            torch.cuda.current_device()
        ).multi_processor_count
    return _SM_COUNT


def k_tiles_for_active_rows(
    _total_active: torch.Tensor,
    k_total_pad: int,
    block_m: int,
) -> int:
    """Second launch dimension for fused sparse MoE kernels (grid ``(T, K_TILES)``)."""
    k_cap = triton.cdiv(k_total_pad, block_m)
    return max(1, k_cap)


def _get_moe_config(w_gate, w2_shape, k_experts, dtype, T):
    config_dtype = _get_config_dtype_str(
        use_fp8_w8a8=False, use_int8_w8a16=False,
        use_int4_w4a16=False, ocp_mx_scheme=None, dtype=dtype,
    )
    return functools.partial(
        try_get_optimal_moe_config, w_gate.size(), w2_shape,
        k_experts, config_dtype, block_shape=None,
    )(T)


# ---------------------------------------------------------------------------
# Triton kernels
# ---------------------------------------------------------------------------

if _SPARSE_MOE_AVAILABLE:
    @triton.jit
    def sparse_moe_threshold_kernel(
        GATE_RAW_ptr,
        TOPK_IDS_ptr,
        FLAT_ROWS_ptr,
        ACTIVE_COUNT_ptr,
        TOTAL_ACTIVE_ptr,
        gate_cutoff,
        N:           tl.constexpr,
        K_EXPERTS:   tl.constexpr,
        K_TOTAL_PAD: tl.constexpr,
        BLOCK_TILE:  tl.constexpr,
    ):
        pid_t = tl.program_id(0)
        flat_offset = 0

        for e in tl.static_range(K_EXPERTS):
            te        = pid_t * K_EXPERTS + e
            gate_base = te * N

            expert_id = tl.load(TOPK_IDS_ptr + pid_t * K_EXPERTS + e)
            flat_base = expert_id * N

            running = 0
            for ts in tl.range(0, N, BLOCK_TILE):
                offs = ts + tl.arange(0, BLOCK_TILE)
                mask = offs < N
                g = tl.load(GATE_RAW_ptr + gate_base + offs,
                            mask=mask, other=0.0).to(tl.float32)

                active = (g >= gate_cutoff) & mask
                slot   = tl.cumsum(active.to(tl.int32))
                count  = tl.sum(active.to(tl.int32))

                wpos  = flat_offset + running + slot - 1
                wmask = active & (wpos < K_TOTAL_PAD)

                tl.store(FLAT_ROWS_ptr + pid_t * K_TOTAL_PAD + wpos,
                         (flat_base + offs).to(tl.int32), mask=wmask)

                running = running + count

            actual = tl.where(running > K_TOTAL_PAD - flat_offset,
                              K_TOTAL_PAD - flat_offset, running)
            tl.store(ACTIVE_COUNT_ptr + te, actual)
            flat_offset = flat_offset + actual

        tl.store(TOTAL_ACTIVE_ptr + pid_t, flat_offset)

    @triton.jit
    def sparse_moe_fused_updown_kernel(
        X_ptr,
        W_UP_ptr,
        W_DOWN_ptr,
        GATE_RAW_ptr,
        TOPK_IDS_ptr,
        TOPK_WEIGHTS_ptr,
        FLAT_ROWS_ptr,
        TOTAL_ACTIVE_ptr,
        PARTIAL_ptr,
        SKIP_STATS_ptr,
        D:           tl.constexpr,
        N:           tl.constexpr,
        K_EXPERTS:   tl.constexpr,
        BLOCK_KE:    tl.constexpr,
        K_TOTAL_PAD: tl.constexpr,
        K_TILES_GRID: tl.constexpr,
        BLOCK_M:     tl.constexpr,
        BLOCK_K:     tl.constexpr,
        BLOCK_N:     tl.constexpr,
        DTYPE:       tl.constexpr,
        RECORD_SKIPS: tl.constexpr,
        stride_xT,   stride_xD,
        stride_wuE,  stride_wuN,  stride_wuD,
        stride_wdE,  stride_wdN,  stride_wdD,
        stride_pT,   stride_pM,
    ):
        pid_t = tl.program_id(0)
        pid_m = tl.program_id(1)

        total_active = tl.load(TOTAL_ACTIVE_ptr + pid_t)
        tile_start = pid_m * BLOCK_M
        if tile_start >= total_active:
            partial_base = PARTIAL_ptr + pid_t * stride_pT + pid_m * stride_pM
            for z in tl.range(0, D, BLOCK_N):
                zoffs = z + tl.arange(0, BLOCK_N)
                zmask = zoffs < D
                tl.store(
                    partial_base + zoffs,
                    tl.zeros((BLOCK_N,), dtype=tl.float32),
                    mask=zmask,
                )
            if RECORD_SKIPS:
                lin_skip = pid_t * K_TILES_GRID + pid_m
                tl.store(SKIP_STATS_ptr + lin_skip, 1)
            return

        if RECORD_SKIPS:
            lin_full = pid_t * K_TILES_GRID + pid_m
            tl.store(SKIP_STATS_ptr + lin_full, 0)

        offs_m = tile_start + tl.arange(0, BLOCK_M)
        mask_m = offs_m < total_active

        fr = tl.load(FLAT_ROWS_ptr + pid_t * K_TOTAL_PAD + offs_m,
                     mask=mask_m, other=0)

        expert_id_m = fr // N
        n_m         = fr  % N

        h_acc = tl.zeros((BLOCK_M,), dtype=tl.float32)
        for k_start in tl.range(0, D, BLOCK_K):
            offs_k = k_start + tl.arange(0, BLOCK_K)
            mask_k = offs_k < D
            x_blk = tl.load(
                X_ptr + pid_t * stride_xT + offs_k * stride_xD,
                mask=mask_k, other=0.0,
            )
            wu = tl.load(
                W_UP_ptr + expert_id_m[:, None] * stride_wuE
                         + n_m[:, None]         * stride_wuN
                         + offs_k[None, :]      * stride_wuD,
                mask=mask_m[:, None] & mask_k[None, :], other=0.0,
            )
            h_acc += tl.reshape(tl.dot(wu, x_blk[:, None]), (BLOCK_M,))

        k_offs  = tl.arange(0, BLOCK_KE)
        k_valid = k_offs < K_EXPERTS
        topk_ids_t = tl.load(TOPK_IDS_ptr     + pid_t * K_EXPERTS + k_offs,
                             mask=k_valid, other=-1)
        topk_w_t   = tl.load(TOPK_WEIGHTS_ptr + pid_t * K_EXPERTS + k_offs,
                             mask=k_valid, other=0.0).to(tl.float32)

        match = (expert_id_m[:, None] == topk_ids_t[None, :])
        rw = tl.sum(match.to(tl.float32) * topk_w_t[None, :], axis=1)
        slot_m = tl.sum(match.to(tl.int32) * k_offs[None, :], axis=1)

        gate_val = tl.load(
            GATE_RAW_ptr + (pid_t * K_EXPERTS + slot_m) * N + n_m,
            mask=mask_m, other=0.0,
        ).to(tl.float32)
        ag = gate_val * tl.sigmoid(gate_val)

        h = rw * ag * h_acc

        h_row = tl.reshape(h.to(DTYPE), (1, BLOCK_M))
        partial_base = PARTIAL_ptr + pid_t * stride_pT + pid_m * stride_pM

        for n_start in tl.range(0, D, BLOCK_N):
            offs_n = n_start + tl.arange(0, BLOCK_N)
            mask_n = offs_n < D
            wd = tl.load(
                W_DOWN_ptr + expert_id_m[:, None] * stride_wdE
                           + n_m[:, None]         * stride_wdN
                           + offs_n[None, :]      * stride_wdD,
                mask=mask_m[:, None] & mask_n[None, :], other=0.0,
            )
            contrib = tl.reshape(tl.dot(h_row, wd), (BLOCK_N,))
            tl.store(partial_base + offs_n, contrib, mask=mask_n)

    @triton.jit
    def sparse_moe_accumulate_kernel(
        PARTIAL_ptr,
        OUT_PTR,
        K_TILES,
        D:      tl.constexpr,
        BLOCK_N: tl.constexpr,
        DTYPE:  tl.constexpr,
        stride_pT, stride_pM,
        stride_outT,
    ):
        pid_t = tl.program_id(0)
        pid_n = tl.program_id(1)

        offs_n = pid_n * BLOCK_N + tl.arange(0, BLOCK_N)
        mask_n = offs_n < D

        acc = tl.zeros((BLOCK_N,), dtype=tl.float32)
        for k_tile in tl.range(0, K_TILES):
            part = tl.load(
                PARTIAL_ptr + pid_t * stride_pT + k_tile * stride_pM + offs_n,
                mask=mask_n, other=0.0,
            )
            acc += part

        tl.store(
            OUT_PTR + pid_t * stride_outT + offs_n,
            acc.to(DTYPE),
            mask=mask_n,
        )

# ---------------------------------------------------------------------------
# Workspace management
# ---------------------------------------------------------------------------

class MoEWorkspace:
    """Grow-never-shrink GPU byte buffer."""

    def __init__(self):
        self._buf: torch.Tensor | None = None

    def ensure(self, nbytes: int, device: torch.device) -> torch.Tensor:
        if self._buf is None or self._buf.numel() < nbytes:
            self._buf = torch.empty(nbytes, dtype=torch.uint8, device=device)
        return self._buf

    def clear(self) -> None:
        self._buf = None


_hybrid_workspaces: dict[int, MoEWorkspace] = {}
_partial_workspaces: dict[int, MoEWorkspace] = {}

# Per-layer sparse config: (gate_cutoff from moe_sparse_sigma, k_experts).
# renormalize is accepted in register_sparse_config for call-site compatibility
# but is not stored  vLLM router handles renormalization internally.
_layer_sparse_configs: dict[int, tuple[float, int]] = {}

_w_down_T_cache: dict[int, torch.Tensor] = {}

_skip_prog_total: int = 0
_skip_prog_skipped: int = 0

_skip_dummy_by_device: dict[torch.device, torch.Tensor] = {}


def _get_skip_dummy(device: torch.device) -> torch.Tensor:
    t = _skip_dummy_by_device.get(device)
    if t is None:
        t = torch.zeros(1, dtype=torch.int32, device=device)
        _skip_dummy_by_device[device] = t
    return t


def reset_fused_skip_program_stats() -> None:
    global _skip_prog_total, _skip_prog_skipped
    _skip_prog_total = 0
    _skip_prog_skipped = 0


def fused_skip_program_stats() -> dict[str, float]:
    tot = _skip_prog_total
    sk = _skip_prog_skipped
    return {
        "programs_total": float(tot),
        "programs_skipped": float(sk),
        "skip_fraction": float(sk / tot) if tot else 0.0,
    }


def register_sparse_config(
    layer_id: int, sigma: float, k_experts: int, renormalize: bool,
) -> None:
    """Pre-register per-layer sparse config so the custom_op can look it up."""
    _layer_sparse_configs[layer_id] = (sigma, k_experts)


def _get_workspaces(layer_id: int):
    if layer_id not in _hybrid_workspaces:
        _hybrid_workspaces[layer_id] = MoEWorkspace()
        _partial_workspaces[layer_id] = MoEWorkspace()
    return _hybrid_workspaces[layer_id], _partial_workspaces[layer_id]


def clear_workspaces() -> None:
    for ws in _hybrid_workspaces.values():
        ws.clear()
    for ws in _partial_workspaces.values():
        ws.clear()
    _hybrid_workspaces.clear()
    _partial_workspaces.clear()
    _layer_sparse_configs.clear()
    _w_down_T_cache.clear()
    _skip_dummy_by_device.clear()


# ---------------------------------------------------------------------------
# Forward implementation
# ---------------------------------------------------------------------------

def _sparse_moe_forward_impl(
    hidden_states: torch.Tensor,
    topk_weights: torch.Tensor,   # [T, K] fp32 from vLLM router
    topk_ids: torch.Tensor,       # [T, K] int32 from vLLM router
    w13_weight: torch.Tensor,
    w2_weight: torch.Tensor,
    layer_id: int,
) -> torch.Tensor:
    sigma, k_experts = _layer_sparse_configs[layer_id]
    T, D = hidden_states.shape
    E, N2, _ = w13_weight.shape
    N = N2 // 2
    K = k_experts
    TK = T * K

    w_gate = w13_weight[:, :N, :]
    w_up   = w13_weight[:, N:, :]
    w_down_T = _w_down_T_cache.get(layer_id)
    if w_down_T is None:
        w_down_T = w2_weight.permute(0, 2, 1).contiguous()  # [E, N, D]
        _w_down_T_cache[layer_id] = w_down_T
        # Only offload when sparse-only mode is guaranteed (no dense fallback possible).
        # if os.environ.get("SPARSE_MOE_OFFLOAD_W2", "0") == "1":
        #     w2_weight.data = w2_weight.data.cpu()  # free GPU copy; w_down_T is the live reference

    BLOCK_TILE  = min(1024, _next_pow2(N))
    K_TOTAL_PAD = _estimate_k_total(N, K, sigma)

    device   = hidden_states.device
    dtype    = hidden_states.dtype
    elem     = hidden_states.element_size()
    tl_dtype = tl.bfloat16 if dtype == torch.bfloat16 else tl.float16

    config   = _get_moe_config(w_gate, w2_weight.size(), K, dtype, T)
    ALIGN_BS = config["BLOCK_SIZE_M"]

    ws, partial_ws = _get_workspaces(layer_id)

    _align = 256
    _T_max = max(T, _SPARSE_MAX_T)

    def _al(n): return (n + _align - 1) // _align * _align

    _max_tok_pad_ws = _T_max * K + E * (ALIGN_BS - 1)
    _max_m_blks_ws  = triton.cdiv(_max_tok_pad_ws, ALIGN_BS)

    R1    = _al(_T_max * K * 2)               # topk_weights_ws  fp16
    R2    = _al(_T_max * K * 4)               # topk_ids_ws      int32
    R3    = _al(_T_max * K * N * elem)        # gate_raw
    R4    = _al(_T_max * K_TOTAL_PAD * 4)     # flat_rows        int32
    R6    = _al(_T_max * K * 4)               # active_count     int32
    R6b   = _al(_T_max * 4)                   # total_active     int32
    R_ST  = _al(_max_tok_pad_ws * 4)          # sorted_token_ids int32
    R_EI  = _al(_max_m_blks_ws * 4)           # expert_ids       int32
    R_NTP = _al(4)                            # num_tokens_pp    int32
    R_OUT = _al(_T_max * D * elem)            # out

    total = R1 + R2 + R3 + R4 + R6 + R6b + R_ST + R_EI + R_NTP + R_OUT
    buf   = ws.ensure(total, device)

    off = 0
    o1    = off; off += R1
    o2    = off; off += R2
    o3    = off; off += R3
    o4    = off; off += R4
    o6    = off; off += R6
    o6b   = off; off += R6b
    o_st  = off; off += R_ST
    o_ei  = off; off += R_EI
    o_ntp = off; off += R_NTP
    o_out = off; off += R_OUT

    topk_weights_ws = buf[o1 : o1 + T*K*2     ].view(torch.float16).reshape(T, K)
    topk_ids_ws     = buf[o2 : o2 + T*K*4     ].view(torch.int32  ).reshape(T, K)
    gate_raw        = buf[o3 : o3 + T*K*N*elem].view(dtype        ).reshape(T, K, N)
    flat_rows       = buf[o4 : o4 + T*K_TOTAL_PAD*4].view(torch.int32).reshape(T, K_TOTAL_PAD)
    active_count    = buf[o6 : o6 + TK*4      ].view(torch.int32  ).reshape(TK)
    total_active    = buf[o6b: o6b + T*4       ].view(torch.int32  ).reshape(T)
    _max_tok_pad    = TK + E * (ALIGN_BS - 1)
    _max_m_blks     = triton.cdiv(_max_tok_pad, ALIGN_BS)
    sorted_token_ids = buf[o_st  : o_st  + _max_tok_pad * 4].view(torch.int32)
    expert_ids       = buf[o_ei  : o_ei  + _max_m_blks  * 4].view(torch.int32)
    num_tokens_pp    = buf[o_ntp : o_ntp + 4               ].view(torch.int32)
    out              = buf[o_out : o_out + T*D*elem        ].view(dtype       ).reshape(T, D)

    # Copy vLLM router outputs into fp16 workspace slots expected by Triton kernels
    topk_weights_ws.copy_(topk_weights.to(torch.float16))
    topk_ids_ws.copy_(topk_ids.to(torch.int32))

    # S2b: expert permutation
    ops.moe_align_block_size(
        topk_ids_ws, E, ALIGN_BS,
        sorted_token_ids, expert_ids, num_tokens_pp, None,
    )

    # S3a: gate GEMM
    invoke_fused_moe_triton_kernel(
        hidden_states, w_gate, gate_raw,
        None, None, None,
        sorted_token_ids, expert_ids, num_tokens_pp,
        False, K, config, tl_dtype,
        False, False, False, False, False, None, None,
    )
    gate_raw_flat = gate_raw.reshape(TK, N)

    # S3b: absolute gate cutoff + compact (``sigma`` is ``moe_sparse_sigma``)
    sparse_moe_threshold_kernel[(T,)](
        gate_raw_flat,
        topk_ids_ws,
        flat_rows,
        active_count, total_active,
        sigma,  # gate_cutoff in kernel
        N=N, K_EXPERTS=K, K_TOTAL_PAD=K_TOTAL_PAD,
        BLOCK_TILE=BLOCK_TILE,
    )

    _maybe_print_sparse_map(layer_id, sigma, gate_raw_flat, T, K, N)

    # S4a/b: fused up+down + accumulate
    _T_ref = _SPARSE_MAX_T
    _sm = _sm_count()
    _bm = min(128, max(32, _next_pow2(_T_ref * K_TOTAL_PAD // (4 * max(_sm, 1)))))
    while _T_ref * triton.cdiv(K_TOTAL_PAD, _bm) < _sm and _bm > 16:
        _bm //= 2
    _bn = 64

    K_TILES = k_tiles_for_active_rows(total_active, K_TOTAL_PAD, _bm)
    _part_buf = partial_ws.ensure(_T_max * K_TILES * D * 4, device)
    partial = _part_buf[: T * K_TILES * D * 4].view(torch.float32).reshape(T, K_TILES, D)

    _record_skips = os.environ.get("VLLM_SPARSE_MOE_RECORD_SKIP_STATS", "") == "1"
    if _record_skips:
        skip_program_buf = torch.zeros(T * K_TILES, dtype=torch.int32, device=device)
        skip_stats_ptr = skip_program_buf
    else:
        skip_stats_ptr = _get_skip_dummy(device)

    sparse_moe_fused_updown_kernel[(T, K_TILES)](
        hidden_states, w_up, w_down_T,
        gate_raw_flat, topk_ids_ws, topk_weights_ws,
        flat_rows, total_active,
        partial,
        skip_stats_ptr,
        D=D, N=N, K_EXPERTS=K, BLOCK_KE=_next_pow2(K), K_TOTAL_PAD=K_TOTAL_PAD,
        K_TILES_GRID=K_TILES,
        BLOCK_M=_bm, BLOCK_K=64, BLOCK_N=_bn,
        DTYPE=tl_dtype,
        RECORD_SKIPS=_record_skips,
        stride_xT=hidden_states.stride(0),  stride_xD=hidden_states.stride(1),
        stride_wuE=w_up.stride(0),    stride_wuN=w_up.stride(1),    stride_wuD=w_up.stride(2),
        stride_wdE=w_down_T.stride(0),stride_wdN=w_down_T.stride(1),stride_wdD=w_down_T.stride(2),
        stride_pT=partial.stride(0),        stride_pM=partial.stride(1),
    )

    if _record_skips:
        torch.cuda.synchronize()
        global _skip_prog_total, _skip_prog_skipped
        n_prog = T * K_TILES
        n_skip = int(skip_program_buf.sum().item())
        _skip_prog_total += n_prog
        _skip_prog_skipped += n_skip

    sparse_moe_accumulate_kernel[(T, triton.cdiv(D, _bn))](
        partial, out,
        K_TILES=K_TILES,
        D=D, BLOCK_N=_bn,
        DTYPE=tl_dtype,
        stride_pT=partial.stride(0), stride_pM=partial.stride(1),
        stride_outT=out.stride(0),
    )

    # One-shot capture (no-op unless SPARSE_MOE_CAPTURE_LAYER is set)
    global _capture_done
    if (
        _CAPTURE_LAYER_ID is not None
        and not _capture_done
        and layer_id == _CAPTURE_LAYER_ID
        and T == _CAPTURE_T
    ):
        import pathlib
        save_path = _CAPTURE_PATH or f"sparse_moe_capture_layer{layer_id}.pt"
        pathlib.Path(save_path).parent.mkdir(parents=True, exist_ok=True)
        torch.save(
            {
                "hidden_states": hidden_states.detach().cpu().clone(),
                "topk_weights":  topk_weights.detach().cpu().clone(),
                "topk_ids":      topk_ids.detach().cpu().clone(),
                "w13_weight":    w13_weight.detach().cpu().clone(),
                "w2_weight":     w2_weight.detach().cpu().clone(),
                "layer_id": layer_id,
                "T": T, "D": D, "E": E, "N": N, "K": K,
            },
            save_path,
        )
        logger.info(
            "[sparse_moe] Captured layer_id=%d T=%d D=%d E=%d N=%d K=%d: %s",
            layer_id, T, D, E, N, K, save_path,
        )
        _capture_done = True

    return out


# ---------------------------------------------------------------------------
# torch.library.custom_op registration
# ---------------------------------------------------------------------------

if _SPARSE_MOE_AVAILABLE:
    @torch.library.custom_op(
        "vllm_sparse_moe::hybrid_forward",
        mutates_args=(),
    )
    def sparse_moe_forward(
        hidden_states: torch.Tensor,
        topk_weights: torch.Tensor,
        topk_ids: torch.Tensor,
        w13_weight: torch.Tensor,
        w2_weight: torch.Tensor,
        layer_id: int,
    ) -> torch.Tensor:
        return _sparse_moe_forward_impl(
            hidden_states, topk_weights, topk_ids,
            w13_weight, w2_weight, layer_id,
        )

    @sparse_moe_forward.register_fake
    def _sparse_moe_forward_fake(
        hidden_states, topk_weights, topk_ids,
        w13_weight, w2_weight, layer_id,
    ):
        T, D = hidden_states.shape
        return hidden_states.new_empty(T, D)

else:
    def sparse_moe_forward(*args, **kwargs) -> torch.Tensor:  # type: ignore[misc]
        raise RuntimeError("Gate-sparse MoE kernel not available.")

\end{lstlisting}

\end{document}